# Natural Language Processing in-and-for Design Research

L Siddharth[*a], Lucienne T.M. Blessing[a], Jianxi Luo[a]

[a] Engineering Product Development, Singapore University of Technology and Design, 8, Somapah Road, Singapore – 487372

## Abstract

We review the scholarly contributions that utilise Natural Language Processing (NLP) techniques to support the design process. Using a heuristic approach, we gathered 223 articles that are published in 32 journals within the period 1991-present. We present state-of-the-art NLP in-and-for design research by reviewing these articles according to the type of natural language text sources: internal reports, design concepts, discourse transcripts, technical publications, consumer opinions, and others. Upon summarizing and identifying the gaps in these contributions, we utilise an existing design innovation framework to identify the applications that are currently being supported by NLP. We then propose a few methodological and theoretical directions for future NLP in-and-for design research.

---

[*] Corresponding author.

Email: siddharth_l@mymail.sutd.edu.sg, siddharthl@iitrpr.ac.in, siddharthl.iitrpr.sutd@gmail.com

# 1. Introduction

Several natural language schemes like ontologies [1], controlled natural language descriptions [2], documentation templates [3], argumentation approaches [4], artefact representations [5], process models [6], and function structures [7] etc., have been adopted in design research to envisage, encode, evaluate, and enhance the design process. While these schemes have significantly impacted the development of several knowledge-based applications in design research and practice, it was not until the development of computational (e.g., Graphical Processing Units (GPUs), cloud computing services) and methodological (e.g., NLTK[1], WordNet[2]) infrastructures that these schemes were popularly utilised to process unstructured natural language text data and extract design knowledge from these. These infrastructures have led to the evolution of what is currently understood and recognized as a family of Natural Language Processing (NLP) techniques.

A typical NLP methodology converts a text into a set of tokens such as meaningful terms, phrases, and sentences that are often embedded as vectors for applying these to standard NLP tasks like similarity measurement, topic extraction, clustering, classification, entity recognition, relation extraction, and sentiment analysis etc. These tasks primarily rely upon prescriptive language tools (e.g., Stanford Dependency Parser[3]), lexicon (e.g., ANEW[4]), and descriptive language models (e.g., BERT[5]).

The ability of NLP methodologies to process unstructured text opens several opportunities like topic discovery [8], ontology extraction [9], document structuring [10], search summarisation [11], keyword recommendation [12], text generation [13] etc., which enable design scholars and practitioners to support knowledge reuse [14], needs elicitation [15], biomimicry [16], [17], emotion-driven design [18] etc., in the design process. NLP has therefore become an imperative strand of design research, where the scholars have extensively proposed NLP-based tools, frameworks, and methodologies that are aimed to assist the participants in the design process, who otherwise often rely upon organisational history and personal knowledge to make important decisions, e.g., choosing a lubricant for shaft interface.

---

[1] Natural Language Toolkit - https://www.nltk.org/

[2] http://wordnetweb.princeton.edu/perl/webwn

[3] http://nlp.stanford.edu:8080/parser/

[4] Affective Norms for English Words –
https://pdodds.w3.uvm.edu/teaching/courses/2009-08UVM-300/docs/others/everything/bradley1999a.pdf

[5] Bidirectional Encoder Representations from Transformers – https://github.com/google-research/bert

In this article, we review scholarly contributions that have applied as well as developed NLP techniques to process unstructured natural language text and thereby support the design process. Several motivations have led to the effort of reviewing such contributions.

1. To identify the methodological advancements that are necessary to bolster the performances of future NLP applications in-and-for design. For instance, the performances of Parts-of-Speech (POS) Tagger and Named Entity Recognition (NER) require significant improvement to process design documents. We have listed various possibilities of such methodological directions in Section 4.2.
2. To enhance theoretical understanding of the nature and role of natural language text in the design process. For example, it is still unclear as to which elements of design knowledge are necessary to be present in an artefact description so that it qualifies as adequate. We have asked several open questions along with necessary discussion to highlight such theoretical directions in Section 4.3.
3. To summarize a large body of NLP contributions into a single source. A variety of NLP applications to the design process are reported in journals outside the agreed scope of design research. Reviewing and summarizing such contributions in this article could therefore be of importance. We have reviewed the contributions according to the type of text source in Section 3.
4. To create an NLP guide for developing applications to support the design process. For example, design methods like creating activity diagrams could be significantly benefited by NLP methodologies. We have indicated such cases in Section 4.1 using a design innovation process framework.

In line with the motivations described above, we adopt a heuristic approach (Section 2 and APPENDIX I) to retrieve 223 articles encompassing 32 academic journals. We review these articles in Section 3 according to the types of text sources and discuss these in Section 4 regarding applications and future directions.

## 2. Methodology

To retrieve the articles for our review, we use the Web of Science[6] portal, where we heuristically search the titles, abstracts, and topics using a tentative set of keywords within design journals. Upon carrying out a frequency-based analysis of the preliminary results, we expand the keyword list as well as the set of design journals. We further expand our search to all journals that include NLP contributions to the design process. We then apply several filters and manually read through the titles, abstracts, and full texts of a selected number of articles. In the end, we obtain 223 articles that we review in our work. We detail the search process in APPENDIX I. We have also uploaded the bibliometric data for all these articles on GitHub[7].

As shown in Table 1, the final set of papers is distributed across 32 journals. We have strategically chosen these journals such that these are primarily design-oriented and secondarily focused on general computer applications (e.g., Computers in Industry), artificial intelligence (e.g., Expert Systems with Applications), and technology-related (e.g., World Patent Information). In addition, we have also included journals that focus on general design aspects such as ergonomics, requirements, and safety.

---

[6] https://mjl.clarivate.com/search-results
[7] https://github.com/siddharthl93/nlp_review/blob/4b9e6b378c8df0bbf61a36e466a50dbb5a0a65d2/nlp_review_papers.csv

Table 1: Article count w.r.t., journals.

| # | Journal Name | Count |
|---|---|---|
| 1 | Journal of Mechanical Design * | 34 |
| 2 | Advanced Engineering Informatics | 24 |
| 3 | Artificial Intelligence for Engineering Design Analysis and Manufacturing * | 24 |
| 4 | Journal of Computing and Information Science in Engineering * | 19 |
| 5 | Expert Systems with Applications | 16 |
| 6 | Computers in Industry | 16 |
| 7 | Journal of Engineering Design * | 15 |
| 8 | Research in Engineering Design * | 15 |
| 9 | Design Studies * | 7 |
| 10 | Engineering Applications of Artificial Intelligence | 7 |
| 11 | Knowledge-Based Systems | 6 |
| 12 | Scientometrics | 5 |
| 13 | Requirements Engineering | 3 |
| 14 | Design Science * | 3 |
| 15 | Design Journal * | 2 |
| 16 | Journal of Computational Design and Engineering | 2 |
| 17 | Concurrent Engineering-Research and Applications | 2 |
| 18 | Decision Support Systems | 2 |
| 19 | Codesign-International Journal of Cocreation in Design and the Arts * | 2 |
| 20 | International Journal of Design Creativity and Innovation * | 2 |
| 21 | Applied Ergonomics | 2 |
| 22 | International Journal of Interactive Design and Manufacturing | 2 |
| 23 | Engineering with Computers | 2 |
| 24 | Ergonomics | 2 |
| 25 | Word Patent Information | 2 |
| 26 | Computer-Aided Design * | 1 |
| 27 | Applied Artificial Intelligence | 1 |
| 28 | International Journal of Design * | 1 |
| 29 | International Journal of Technology and Design Education | 1 |
| 30 | Technovation | 1 |
| 31 | Reliability Engineering and System Safety | 1 |
| 32 | Artificial Intelligence in Engineering | 1 |

* Indicates the journals that we initially considered as those that fall within the scope of design.

As shown in the year-wise plot (Figure 1), there has been a steady increase in the number of contributions, which could be mainly due to the evolution of computational and methodological infrastructures[8]. While the contributions from the 1990s have been theoretically influential, the peak in the mid-2000s could be attributed to the popularity of biomimicry [19], ontologies [20], functional modelling [21], and functional representation [22] etc. Besides the year-wise plot, we report the 30 most frequent keywords as a word cloud in Figure 2, where we discard the generic keywords such as 'design', 'system' etc.

---

[8] By infrastructures, we mean the following: a) commercial access to high-end computing systems via services like Amazon, Google Cloud etc., b) open sources Application Programming Interfaces (APIs) provided by several data sources like YouTube, PatentsView etc., c) the accessibility to open sources codes and algorithms via online communities like HuggingFace.

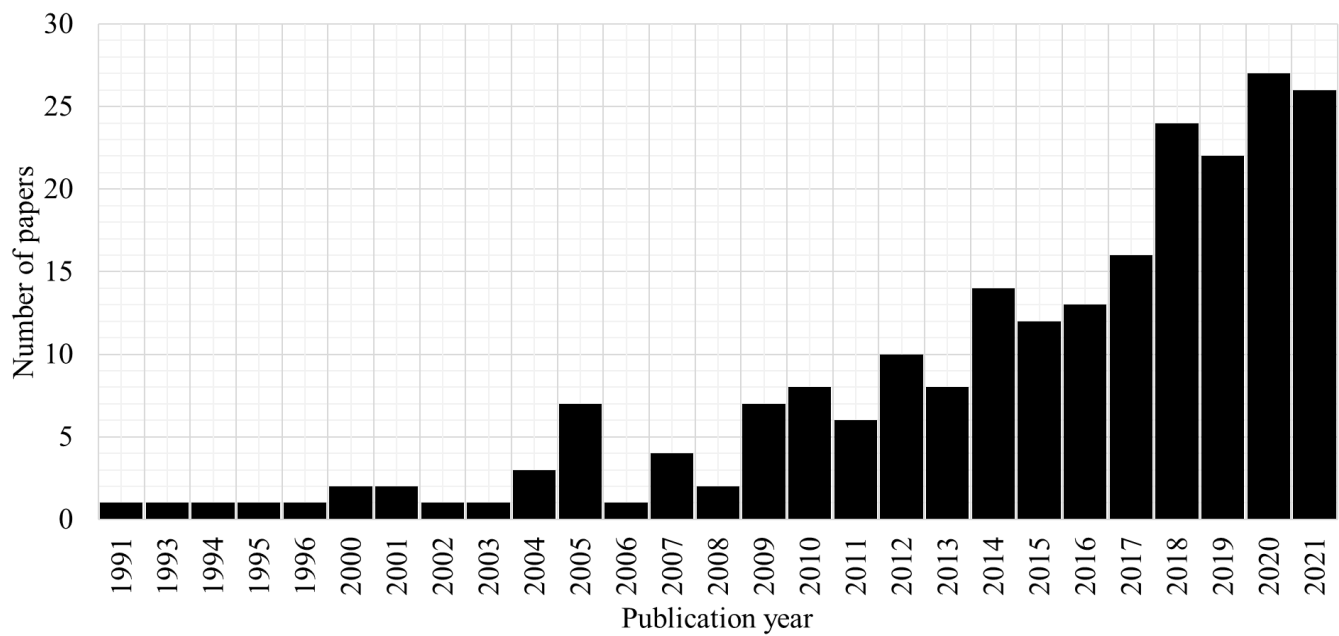

Figure 1: Article count w.r.t., the publication year. The data point at 2021 is applicable only until 19[th] September 2021.

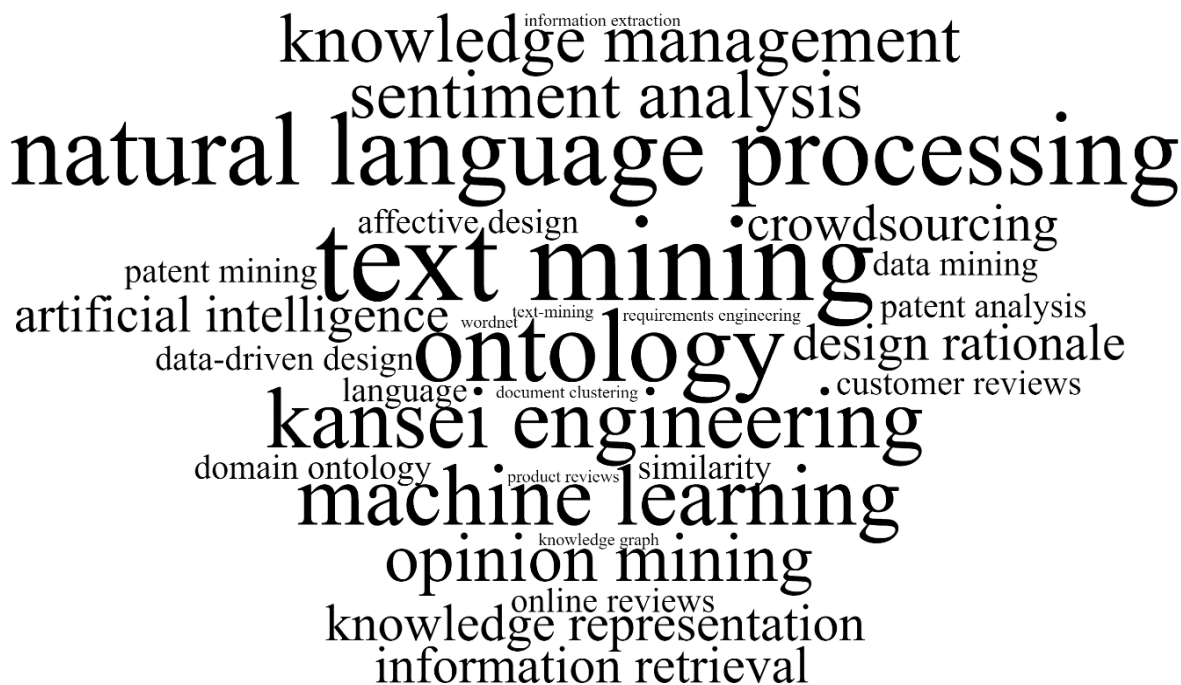

Figure 2: Top 30 keywords w.r.t., frequency.

# 3. Review

In this section, we review the 223 articles[9] thus selected using the methodology as described in Section 2 and APPENDIX I. To present the articles that we have reviewed, we considered the following categorisation schemes: 1) the types of natural language text data, e.g., internal reports, technical publications etc., 2) the types of NLP tasks, e.g., clustering, classification etc., and 3) the applications in the design process, e.g., brainstorming, problem formulation etc. Among these schemes, we adopt the types of text data because an NLP-based contribution is often associated with one text source data but combines a variety of NLP tasks and could be applied across different phases of the design process.

As shown in Figure 3, we map the categories of our scheme onto different phases of the design process as given in the model of the UK Design Business Council[10]. Among the types of text data sources as explained below, consumer opinions and technical publications are utilised in the design process, while the rest are generated in the design process.

- The **internal reports** are usually generated in the *deliver* phase of the design process, where the concepts are embodied and detailed into prototypes. These sources of natural language text often include the knowledge of failures, situations, logs, instructions etc.
- The **design concepts** are generated during the *develop* phase, when the designers search, retrieve, associate, and select concepts using various supports. The NLP contributions that we review under this category not only involve processing design concepts but also problem statements, keywords, supporting databases (e.g., AskNature) etc.
- The **discourse transcripts** constitute the recorded communication like speech transcripts, emails etc., that are obtained from organisational data or think-aloud experiments. These sources need not capture the communication that is pertinent to a particular phase but the design process as a whole.
- The **technical publications** that constitute patents, scientific articles, and textbooks are considered external sources that are often utilised in the *develop* and *deliver* phases of the design process. Owing to the quality and quantity of text, these sources are best suited for the application of NLP tasks.

---

[9] Wherever applicable, we also review the supporting and relevant articles alongside the selected 223 articles.

[10] We chose the double diamond model for representing the design process because of its diversity across different streams of design research. If we were to utilise comprehensive, yet specific models, e.g., Pahl and Beitz [23] that was utilised by Chiarello et al. [24], it is difficult to choose between embodiment and detailed design phases for categorising articles.

- The **consumer opinions** are external sources that are available in the form of product reviews and social media posts. These sources are predominantly utilized in the *discover* phase of the design process, when the designers understand the usage scenarios and extract user needs.
- We categorise the miscellaneous contributions as '**other sources**' that are not indicated in Figure 3.

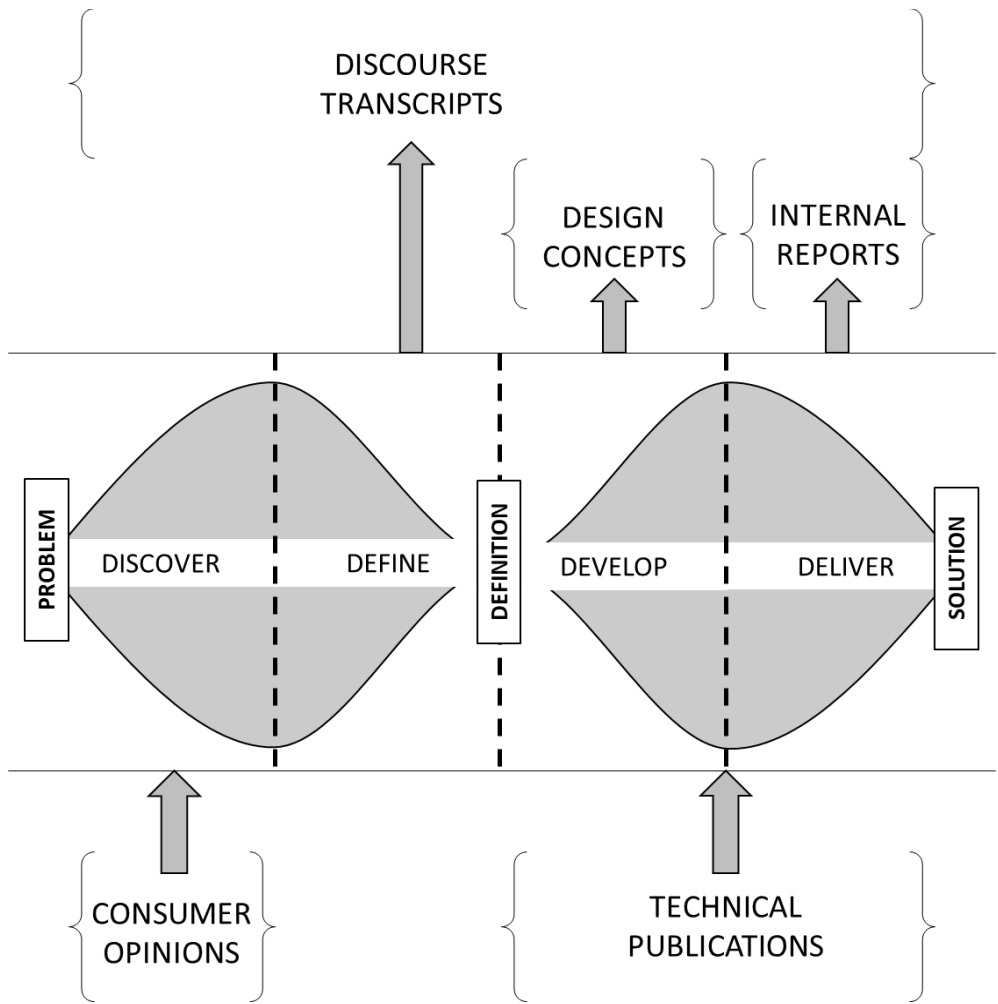

Figure 3: Indication of natural language text data sources related to the design process, following the 4D design process model from the UK Design Business Council.

## 3.1. Internal Reports

Internal reports constitute over 80% of the knowledge in the industry [25] and are often present as product specifications, design rationale, design reports, drawing notes, and logbooks [26]. Although conventional NLP methodologies like building classifiers [27] using internal reports are a recent phenomenon in design research, scholars have attempted to process internal reports and discover ontologies [28] since the early '90s.

*3.1.1. Requirement Extraction*

Scholars initially aimed to extract design requirements as meaningful terms, phrases, and segments from internal reports to reuse these in the design process. Such requirements shall also be derived from the past cases of failure in which violated constraints were recorded [29]. As mentioned below, scholars initially encountered some challenges while extracting design requirements from internal reports.

Kott and Peasant [30, p. 94] observe that requirements in internal reports are incomplete, ambiguous, include inconsistent rationale and denote a wrong purpose. To mitigate some of these issues, they provide an example [30, p. 103] as shown below to illustrate how lengthy requirements could be decomposed into short sentences.

> *"The Loader shall provide the capability of handling HCU6/E pallets, ISO 40-foot containers, and Type V airdrop platforms. Loader shall be able to move forward with speed of at least 5 mph, the goal being 7 mph. An on-board maintenance diagnostic system shall be provided."*

> *"The Loader shall be able to perform the Loading function. The Load Type of the Loading function shall be any of: HCU6/E pallets, ISO 40-foot containers, and Type V airdrop platforms. The Loader shall be able to perform function Move Forward. The speed of Move Forward shall be at least 5 mph, the goal being 7 mph. The Loader shall include an On-Board Maintenance Diagnostic System."*

Farley [31, pp. 296, 299] identifies that airtime faults (also called 'snags') include abbreviations (e.g., CHKD – checked, S0V – serviceable), acronyms, spelling errors (e.g., VLVE), and plural terms. While differing in structure and semantics [32, p. 155], internal reports also include noisy terms [33, p. 179], 'plastic' terms (e.g., 'progress', 'planning'), and implicit phrases (e.g., "insufficient performance") [34, p. 62]. Kim et al. [32, p. 162] suggest that acronyms ('CNC') and abbreviations ('chkd') shall be recognized in text using ontologies. To reduce ambiguity, Madhusudhanan et al. [35, p. 451] suggest that the anaphora ('those') shall be replaced with the corresponding entity in the previous sentence.

When co-ordination ambiguity exists in a sentence, for example, "slot widths and radii should conform to those of cutters" [36, p. 2], it is unclear if the term 'slot' modifies 'widths' or 'radii'. Here, Kang et al. [36, pp. 6, 7] suggest checking if the corresponding domain ontology includes ('slot', 'hasProperty', 'radii'). To extract meaningful segments that are devoid of ambiguities, Madhusudanan et al. [35, p. 452] measure coherence between sentences by integrating and extending WordNet-based similarity measures. To extract segments within a sentence, for example, "*sharp corners should be avoided* because they interfere with the metal flow", Kang et al. [37, p. 294] extract the *italicized* portion using domain concepts (e.g., corner) and attributes (e.g., isSharp). They also discard the unwanted portion using some rules [37, p. 295]; e.g., the subordinate clause that occurs after a marker shall be discarded, except for 'if' or 'unless'.

*3.1.2. Ontology Construction*

To represent design rationale[11], scholars have proposed a variety of prescriptive-generic ontologies [38]–[42] that build upon the fundamental idea of entity-relationship models [43]. While generic ontologies are capable of capturing rationale from a variety of domains, the performances of these in terms of knowledge retrieval are expected to be low due to the level of abstraction. For example, a list of generic terms that represent 'issue' [40, p. 4] may not retrieve phrases that inherently or intricately communicate a design issue.

Domain-specific ontologies like QuenchML [44] and Kodak Cover [45] overcome the limitations of generic ontologies while also being evolvable [46], machine-readable [47], [48], and semantically interoperable [49]. Scholars have therefore attempted to extract domain-specific ontologies from domain text sources.

Among domain-specific ontologies, Kim et al. [32, p. 160] identify the following categories of relationships from aircraft engine repair notes: background, cause-effect, condition, contrast, etc. Lough et al. [34, p. 33] understand from 117 risk statements that these are indicators of failure modes, performance, design, and noise parameters. Using oil platform accident reports, Garcia et al. [50, pp. 430, 431] propose that concept relationships could be generalized as Is-a, Part-Of, Is-an-attribute-of, Causes, Time-Follows, Space-Follows and more. Hsiao et al. [51, p. 147] populate 822 actions contained in 185 risk reports and identify that action could carry the attributes 'purpose' and 'embodiment', which are further categorised as 'Approval', 'Gather_data', 'Coordinate', 'Request' etc., [51, p. 158].

Scholars have built ontologies by associating technical terms and segments using various similarity measures. Hiekata et al. [52] use an existing ontology to associate word segments (component and malfunction) from 9604 shipyard surveyor reports. Lee et al. [3] mine the task data from shipbuilding transportation logs and cluster these using a variety of distances (e.g., Jaccard, Euclidean). Kang and Tucker [53] extract functions as topic vectors from 16 module descriptions [54] of an automotive control system. They propose that the cosine similarity between a pair of topic-vectors (function) quantifies the functional interaction between corresponding modules. Song et al. [55, pp. 265–269] construct a semantic network using iPhone Apps Plus[12] text data that includes 697 service documents indicating 66 feature elements and 95 feature keywords.

---

[11] Design rationale that is mentioned here refers to the previously recorded decisions and underlying reasons that are pertinent to past design issues.

Arnarsson et al. [56] use Latent Dirichlet Allocation (LDA) to cluster the Doc2Vec-based embeddings of over 8,000 Engineering Change Requests (ECRs) in a commercial vehicle manufacturer. Yang et al. [57] construct an ontology using 114,793 problem-solution records within pre-assembly reports inside an automotive manufacturer. They use the ontology to process (e.g., identify n-grams), structure, and represent new text data in various forms [57, p. 214] to facilitate the design and managerial decisions. Xu et al. [58] obtain the text data of 1844 problems and 1927 short-term remedies from a vehicle manufacturer. To link the problems and remedies, they transform the text using Term Frequency - Inverse Document Frequency (TD-IDF) and perform K-means clustering for problems and short-term remedies, while also linking the clusters.

*3.1.3. Design Knowledge Retrieval*

While "knowledge retrieval" could assume a broad meaning across different areas of research and practice, we mention this in reference to the methods that 'retrieve' terms, phrases and segments that include components, issues, constraints, interactions etc., The outputs of such retrieval methods shall be considered as 'design knowledge' if it is possible to re-represent these as <entity, relationship, entity> triples that form constituents of an artefact that is relevant to the design process.

For example, a segment extracted from a transistor patent [59, p. 8] – "an insulating material is deposited on the whole surface of the substrate having the first semiconductor layer" shall be encoded into triples such as <insulating material, is deposited, whole surface>, <whole surface, of, substrate>, and <substrate, having, first semi-conducted layer> that form the constituents of the patent that shall be utilised as knowledge aid in the design process. To extract such relevant terms, phrases, and segments, the scholars have adopted a couple of directions. First, using an ontology that shares the same domain as target text data so that relevant portions of the text are identified. Second, indexing the unstructured text data using a classification algorithm so that the search is restricted to the relevant portions.

---

[12] http://www.iphoneappsplus.com/

To assist case-based reasoning, Guo et al. [60] build a domain ontology using 1000 injection moulding cases that were encountered in a Shenzhen-based company. They demonstrate using an Information-Content (IC) based similarity measure as to how the ontology aids in knowledge retrieval. For case-based retrieval, Akmal et al. [61] compare a variety of ontology-based similarity measures (e.g., Tversky's Index, Dice's Co-efficient) against numeric similarity measures (e.g., Wu-Palmer, Lin) to observe that the former deviated less from expert's similarity scores. To retrieve CAD models using text inputs, Jeon et al. [62] demonstrate how ontologies could be used as intermediaries. To assist CAD designers with design rule recommendations, Huet et al. [63] create a knowledge graph around a design rule using relationships such as 'has keyword' (semantic context), 'has material' (engineering context), 'has employee' (social context) etc.

To relate phenomena and failure modes, Wang et al. [64] extract a lightweight ontology from 400 aviation engine failure analysis reports and utilise the ontology to represent phenomena and failure modes as attribute-value vectors. They [64, pp. 270, 271] then map the phenomena and failure mode vectors using an artificial neural network. To extract candidate components and responsibilities from the design rationale text, Casamayor et al. [65] obtain sentences from IBM supported rationale suite[13] and the UNICEN university repository[14]. Upon classifying the sentences as functional or non-functional using a semi-supervised approach, they extract verb phrases as candidate responsibilities and group these using the hierarchical clustering method to identify candidate components.

To understand the coupling between design requirements, Morkos et al. [10, p. 142] construct a bipartite network of terms and 374 requirements obtained from Toho (160) and Pierburg (214) manufacturing projects. They label a portion of these terms as 'useful' or 'not useful' and vectorize these using the network properties (29 features) and string length (1 feature). They train a neural network using the labelled dataset to classify the rest of the terms. Using the set of terms that are classified as 'useful', they reconstruct the bipartite network and retrain the classifier until the length of the list of terms is saturated [10, p. 149].

To classify and index airtime faults, Tanguy et al. [66] train a Support Vector Machines (SVM) classifier on 136,861 labelled documents that were obtained from the French Aviation Regulator – DGAC. To classify the causes of automotive issues, Xu et al. [67] obtain titles and descriptions of 2,420 issues from a Chinese automotive manufacturer. They retrieve *cause-related* phrases using a domain ontology and label these with the categories of the Fishbone diagram – Man, Machine, Material, Method, and Environment. They use the labelled dataset to train a binary-tree-based SVM classifier.

---

[13] http://www-01.ibm.com/software/rational/

[14] http://isistan.exa.unicen.edu.ar

To identify computer-supported collaborative technologies, Brisco et al. [68, p. 65] obtain Global Design Project text data from 104 students and classify the sentences into requirements, technologies, and technology functionalities using RapidMinerStudio[15]. Lester et al. [69, pp. 133–135] classify the chrome bug reports[16] into requirements, decisions, alternatives etc., using the Naïve Bayes algorithm to find that feature selected using optimization approaches (e.g., Ant-colony) result in higher F-1 measure compared to document characteristics (e.g., TF-IDF).

To index manufacturing rules, Ye and Lu [70] train a feedforward neural network with two hidden layers (128 and 32 neurons) using the embeddings of manufacturing rules and eight category labels. Song et al. [71] train a Bi-directional LSTM using 350 building regulation sentences to extract predicates and arguments. For example, in a design rule – "The roof height of the building must be 15 meters or less," the predicate is "be less" and the arguments are 'roof height', 'building', and '15 meters.' To automatically extract design requirements, Fantoni et al. [72] process tender documents of Hitachi Railway using a variety of ontologies and classify a sentence as a requirement if it includes certain keywords.

*3.1.4. Summary*

We summarize the NLP methodologies applied to Internal Reports in Table 2 according to data, methods, and supporting materials. We indicate the future possibilities of these in red colour. We use the same table format to summarize the literature review for the remaining types of data sources as well. Internal reports mainly include issues and remedies that are pertinent to a specific organisation or a domain. Scholars have used a variety of internal reports to process, extract ontologies, and classify sentences. They have also demonstrated how ontologies are used for effective knowledge retrieval.

---

[15] https://rapidminer.com/products/studio/

[16] Bugs reported on Chrome web browser from the Mining Software Repositories Mining Challenge – http://2011.msrconf.org/msrchallenge.html

Table 2: Summary of NLP methodologies and future possibilities with Internal Reports.

| | Data | Methods | Supporting Materials |
|---|---|---|---|
| **Requirement Extraction** | Airtime Faults (Snags), Manufacturing Rules, Aircraft Assembly Documents, **Technical Discussion Platforms**: e.g., Stack Overflow, Reddit | **Text Segmentation**: Manual, Rule-based Approach, Semantic Similarity (e.g., Lin) **Ambiguity Resolution**: Ontology-based Approach, Rule-based Approach **Term Identification**: Ontology-based Approach | **Lexicon**: WordNet, Generic Technical Lexicon **Ontologies**: Manufacturing |
| **Ontology Construction** | Aircraft Engine Repair Notes, Oil Platform Accident Reports, Shipyard Surveyor Reports, Shipbuilding Transportation Logs, Automotive Control System Descriptions, iPhone service documents, Vehicle Engineering Change Requests, Automotive Pre-Assembly Reports, Vehicle Manufacturing Problems **Other Streams**: Integrated-Circuit Design, Software Interface Design, Algorithm Design, Virtual Spaces | **Embedding**: Doc2Vec, TF-IDF transformation, Domain-specific Language Model **Similarity Measurement**: Cosine, Semantic Similarity measures (e.g., Wu-Palmer) **Clustering**: Distance-based (e.g., Jaccard) method, K-means **Topic Extraction**: Latent Dirichlet Allocation (LDA) **Named Entity Recognition:** Domain-specific Language Model | **Lexicon**: WordNet, Generic Technical Lexicon **Ontologies**: QuenchML, Kodak family, Shipyard, Automotive |
| **Design Knowledge Retrieval** | CAD rules, Aviation Engine Failures, Rationale Suite (IBM), UNICEN repository, Toho Project, Pierburg Project, DGAC Airtime Faults, Automotive Manufacturing Issues, Global Design Project, Chrome Bug Reports, Building Regulation Sentences, Hitachi Railway Tender Documents | **Embedding**: Attribute-Value (using ontology), Bipartite Network Properties, Word2Vec, Domain-specific Language Model **Similarity Measurement:** Information-Content based measures, Ontology-based measures (e.g., Tversky's Index) **Feature Selection**: TF-IDF, Mutual Information, Information Gain, Ant Colony Optimization, Genetic Algorithms, Domain-specific Language Model, Sentence Embedding Models (Doc2Vec) **Clustering**: Hierarchical Clustering **Classification**: Artificial Neural Network (ANN), Rule-based Approach, Feedforward Neural Network, Support Vector Machines (SVM), Binary Tree-based SVM, Rapid Miner, Naïve Bayes, Bi-directional LSTM | **Ontologies**: Injection Moulding, Aviation Engine, Automotive, Railway, Building |

We can observe from the data column of Table 2 that internal reports have been utilised from a variety of domains: Aerospace, Shipbuilding, Automotive etc. Surprisingly, none of the methodologies has utilised data sources from the most popular 'silicon-based streams such as Integrated Circuits, Software Architecture, and Data Structures. While discussion platforms like Stack Overflow and Reddit may not be classified as those included within internal reports, they include the knowledge of issues and solutions that are found both in the industry and academia.

Upon training nearly 0.8 million labelled patent documents for a classification task, Jiang et al. [73] observe that the accuracy tends to be higher when the input feature vectors integrate text, image, and meta information of the document compared to only-text and only-image feature vectors. Hence, the analysis of mere text in *multi-modal* documents like transportation logs [3] may not reflect the entire design knowledge that is being communicated in these.

As understood from the list of data sources, the *accessibility* to internal reports is highly restricted. Although analyses of internal reports have a high probability of extracting design knowledge, these sources are also characterized by low *information content*, e.g., 350 building regulation sentences [71]. While this caveat limits the performances of classifiers, the ontologies extracted from these also may not be comprehensive. Hence, it is necessary to aggregate various internal reports from a domain into a single source of natural language text, e.g., NASA Memorandum on Space Mechanisms – lessons learnt [74].

As far as the methods are concerned, although scholars have applied several state-of-the-art methods to perform NLP tasks such as classification and clustering, they are yet to utilise language models like BERT. While such language models are expected to perform poorly on domain documents [72], it would be significantly useful to develop domain-specific language models, e.g., BioBERT [75]. These models shall be useful for obtaining embeddings and subsequently tasks such as Named Entity Recognition, Text Classification etc. Apart from the cost and resource limitations, training these language models also requires high amounts of text data that does not seem currently feasible with internal reports.

Term identification is a fundamental NLP problem that has not been given enough attention by design scholars apart from those that have utilised internal reports. The terms like "roller bearing" reduce the ambiguity caused by individual words 'roller' and 'bearing'. Since meaningful terms are made of two or more words [76], [77], it is critically important to identify these before applying higher-level NLP tasks. Scholars have resorted to ontology-based approaches to identify these terms [57], [72]. While ontology-based approaches are recommended over common-sense lexicon (e.g., WordNet), it is necessary to rely on domain-specific language models and generic- design- and technical- oriented lexicon to identify general terms (e.g., rough surface). Although such supports are hard to build, there has been recent progress in the literature that adopts patent databases to develop a generic lexicon [78], [79].

Term disambiguation is another fundamental NLP problem, e.g., the terms such as "cathodic protection anode bed", "deep anode well", and "deep ground bed" are often used to refer to "cathodic protection well". [80, p. 5]. Since the ambiguity posed by these terms is concerning the underlying meaning, the approach to resolve this issue should concern the measurement of semantic similarities among these terms. Gu et al. [81, p. 108] resolve semantic conflicts between similar sentences, e.g., "I will buy a bike" and "I will buy a bicycle" using a WordNet-based ontology – FloDL. Such a type of semantic conflict resolution is hardly relevant to industrial applications. While usage of a common-sense lexicon like WordNet is not recommended for such tasks, it is necessary to obtain true embeddings of these terms using domain-specific language models so that cosine similarity reflects 'nearly' actual similarity.

## 3.2. Design Concepts

Often associated with the *develop* phase of the design process (as indicated in Figure 3), design concepts are generated through search, retrieval, association, and selection. The NLP methodologies applied to these stages need not use only concept descriptions as primary text sources but also the problems, keywords, source of stimuli etc.

*3.2.1. Concept Search*

To formulate a comprehensive set of keywords to search for concepts, researchers have sought WordNet for identifying troponyms ('prevent' → 'inhibit') [82, p. 3], [83, pp. 3, 4], bridge verbs [84, p. 50], semantically distant verbs [85, pp. 272, 291], and morphological nouns [86, p. 5]. Chakrabarti et al. [2, pp. 119–121] provide a systematic approach to searching and retrieving biological stimuli based on the SAPPhIRE[17] model. In the Action construct of the SAPPhIRE model, for instance, they propose that the search could be a combination of verb, noun, and adjective. Rosa et al. [87] build upon the approach of Chakrabarti et al. [2] by combining SAPPhIRE and Function-Behaviour-Structure to form a unified ontology for biomimicry.

To effectively search for concepts of biological species, Rosa et al. [88] develop a structured database of these and group these using high-level functions that are represented using <verb, noun, predicate> where the predicate is represented as <preposition, noun>. Vandevenne et al. [89, pp. 21, 22] use k-Nearest Neighbours (k-NN) algorithm to index the AskNature[18] database by classifying 1531 unique analogical transfer strategies into the following levels [89, p. 25]: group (e.g., move or stay put, modify), subgroup (e.g., attach, adapt), and function (e.g., temperature, compression). Chen et al. [90] examine 20 AskNature pages to extract meaningful keywords and structure-function knowledge using, respectively, TF-IDF values and selected dependency patterns.

The above-stated contributions aim to enhance the concept search w.r.t., the biological domain. We also review some approaches that sought other domains. To improve the quality of concepts generated by architects [91], De Vries et al. [92] integrate WordNet-based word graphs and a sketching canvas. To assist novice designers to form domain-specific keywords, Lin et al. [15, p. 356] map user needs and domain concepts through the so-called 'OntoPassages' that were extracted using a domain ontology, which was built using 111 documents belonging to the National Center for Research on Earthquake Engineering, Taiwan.

---

[17] SAPPhIRE is a model of causality that comprises the following constructs: States, Actions, Parts, Phenomena, Inputs, oRgans, and Effects.

[18] https://asknature.org/

To recommend a suitable design method for a problem description, Fuge et al. [93] obtain 886 case study descriptions and method labels from Human-Centred Design (HCD) Connect. They use Latent Semantic Analysis (LSA) to obtain the vectors of the descriptions and train the following classifiers using the labelled dataset: Random Forest, SVM, Logistic Regression, and Naïve Bayes. To enhance problem definition, Chen and Krishnamurthy [94] facilitate human-AI collaboration in completing problem formulation mind maps with the help of ConceptNet and the underlying relationships.

*3.2.2. Concept Retrieval*

The contributions in this section are primarily retrieval systems that are built under the assumption that the problem is well-defined, and the search keywords are known beforehand. Chou [95] and Yan et al. [96] adopt the Su-field problem modelling approach to systematically obtain ideas through TRIZ and manually evaluate these using a fuzzy-linguistic scale. Kim and Lee [97] integrate various design-by-analogy approaches into an interface called Bionic MIR that allows retrieval of biological systems based on physical, biological, and ecological relations.

In a tool named Retriever, for a search keyword (e.g., chair) and a relation (e.g., "is used for") from ConceptNet[19] categories, Han et al. [98, pp. 467–469] retrieve three re-representations (e.g., chair, bench, sofa) and corresponding entities (e.g., leading a meeting, growing plants, reading a book) that are connected by the selected relation. In another tool named Combinator, for the same inputs, Han et al. [99, p. 12/34] retrieve the related entity (noun, verb, adjective) and concatenate it with the search keyword, e.g., 'Handbag' → 'Origami Handbag'.

To support the rapid retrieval of concepts, Goucher-Lambert et al. [100] employ LSA on a design corpus to identify a near or far concept from the current concept. They then provide the concept thus identified from the corpus as a stimulus to the designers for generating more concepts. To demonstrate retrieval of concepts using the C-K theory [101], Li et al. [102] extract a healthcare knowledge graph by mining SVO tripes of the form: $\underbrace{NP_{sub}}_{amod} \xrightarrow{nsubj} \underbrace{VP}_{advmod} \xrightarrow{dobj} \underbrace{NP_{obj}}_{amod}$ from 18,000 Chinese websites. They also populate an FBS-based 'nursing bed' knowledge graph using experts.

---

[19] https://conceptnet.io/

*3.2.3. Concept Association*

Once the concepts are generated using the search and retrieval methods, it is necessary to group similar concepts, especially when a large number of concepts are crowdsourced. In this section, we review NLP contributions that associate concepts predominantly using graph-based approaches. Zhang et al. [103, p. 2] group 930 concepts (described as paragraphs) that were obtained from a human-centred design course[20] using Word2Vec and the hierarchical clustering algorithm. Ahmed and Fuge [104, p. 11,12/30] measure topic level association for 3918 ideas that were submitted to OpenIDEO[21] using a Topic Bison Measure, which indicates if a topic pair co-occurs in an idea as well as the proportions of the pair.

To examine the effectiveness of crowdsourced stimuli, Goucher-Lambert and Cagan [105] crowdsource concepts as three nouns and three verbs for 12 design problems and categorise these as near, far, and medium stimuli based on the frequency and WordNet-based path similarity. He et al. [106] crowdsource text descriptions of thousands of ideas to future transportation systems via Amazon Mechanical Turk. They [106, pp. 3, 4] form a coword network of these ideas and use MINRES[22] to extract core ideas from the network. Liu et al. [107, p. 6] summarize 1,757 scientific articles (solutions to a transmission problem) by building Word2Vec-based semantic networks around the central keywords – {transmission, line, location, measurement, sensor, and wave}. Camburn et al. [108], [109] utilise HDBSCAN [23] for clustering crowdsourced concepts and TextRazor[24] for extracting entities and topics from these.

*3.2.4. Concept Selection*

In this section, we review the contributions that have utilised NLP supports to evaluate and select concepts. These concepts primarily aim to measure one or more success metrics (e.g., novelty). Delin et al. [110, pp.

---

[20] Taught at the University of California, Berkeley

[21] https://www.openideo.com/

[22] https://web.stanford.edu/group/SOL/software/minres/

[23] https://hdbscan.readthedocs.io/en/latest/how_hdbscan_works.html

[24] https://www.textrazor.com/

125–129] use bipolar adjectives obtained from the British National Corpus (BNC) to rate concepts. Strug and Slusarczyk [111] evaluate floor plan concepts using the frequently occurring patterns in the hypergraph representations of past floor plans. To understand the concept selection phenomenon, Dong et al. [112] model the change of linguistic preferences using the Markov Process and calculate the transition probabilities. To calculate creativity, Gosnell and Miller [113, pp. 4–6] tie 27 concepts with some adjectives and match these against the terms – 'innovation' and 'feasibility' using DISCO[25].

Chang and Chen [114] obtain 108 ideas for future personal computers from DesignBoom[26] and mine the idea-related information using RapidMiner. They apply K-means clustering to group the ideas and Analytic Hierarchy Comparison to evaluate these. Siddharth et al. [41, pp. 3–5] measure the novelty of a concept by comparing it against all entries in a reference product database across SAPPhIRE constructs and using a WordNet-based similarity. To examine the success of ideas that were submitted to Kickstarter – a crowdfunding platform, Lee and Sohn [115] shortlist 595 ideas in the Software-Technology category. They apply LDA to extract the most important 50 topics from the text descriptions of these ideas. Using the 50 topics and the funding received by the ideas, they conduct a conjoint analysis to examine the contribution of a topic to the success of an idea [115, pp. 107, 108].

*3.2.5. Summary*

As we have summarized in Table 3, the NLP contributions that are pertinent to Design Concepts assist concept search, retrieval, association, and selection. Scholars have utilised a variety of knowledge bases to search and retrieve concepts, while also recommending novel ways to expand search keywords. Since crowdsourcing concepts have recently emerged as an alternative to traditional laboratory-based design studies, scholars have therefore found the need to associate and group the concepts for assessing these. The NLP applications to concept selection are still emerging as there exist many metrics and many ways to compute these.

---

[25] https://github.com/linguatools/disco/blob/master/src/main/java/de/linguatools/disco/DISCO.java
[26] https://www.designboom.com/

Table 3: Summary of NLP methodologies and future possibilities with Design Concepts.

|  | Data | Methods | Supporting Material |
|---|---|---|---|
| **Concept Search** | Idea-Inspire, AskNature, National Centre for Research on Earthquake Engineering, HCD Connect Case Studies, TRIZ, Encyclopaedia, How Stuff Works, YouTube | **Term Retrieval**: Lexical Relationships, Semantic Similarity, Dependency Parsing, TF-IDF Values, Word Graphs, Mind Maps, Design Knowledge Graph<br>**Embedding**: Latent Semantic Analysis (LSA), BERT, GPT-x<br>**Classification**: k - Nearest Neighbours (k-NN), Random Forest, SVM, Logistic Regression, Naïve Bayes, LSTM variants | **Lexicon**: WordNet, ConceptNet<br><br>**Ontology**: SAPPhIRE, FBS, Earthquake |
| **Concept Retrieval** | Bionic MIR, Healthcare Websites, Nursing Bed Knowledge Graph, YouTube, Google Patents | **Term Retrieval**: Semantic Similarity, Subject-Verb-Object Triples, Source Domain Ontologies, Google API<br>**Similarity Measurement**: LSA, BERT, GPT-x | **Lexicon**: WordNet, ConceptNet<br>**Ontology**: C-K theory, FBS, SAPPhIRE |
| **Concept Association** | Human-Centred Design Course, OpenIDEO, Crowdsourced Transportation Concepts, Scientific Articles (Transmission Problem) | **Similarity Measurement**: Word2Vec, Topic Bison Measure, Path Similarity<br>**Embedding**: BERT, GPT-x, Domain-specific language model<br>**Clustering**: Hierarchical Clustering, HDBSCAN<br>**Topic Extraction**: MINRES, TextRazor | **Lexicon**: WordNet |
| **Concept Selection** | Floor Plan Concepts, DesignBoom, Kickstarter, Red Dot Design Awards, Malcolm Baldrige National Quality Award, Award Patents, Nielsen Retail Scanner Data, Standard Datasets | **Similarity Measurement**: Hypergraph Pattern Matching, DISCO, Analytic Hierarchy Comparison, WordNet-based Similarity<br>**Term Retrieval**: RapidMiner<br>**Clustering**: K-means<br>**Topic Extraction**: LDA<br>**Others**: Markov Process, Conjoint Analysis | **Lexicon**: British National Corpus, WordNet, Affective Lexicon<br>**Ontology**: SAPPhIRE, FBS |

While scholars have utilised both general and domain-specific text sources for searching concepts, it is also possible to explore more text sources such as Encyclopaedia and How and Stuff Works. One of the most consulted platforms – YouTube seems unexplored. Although being primarily a video-sharing platform, the descriptions, comments, and captions on YouTube are still useful text sources for inspiration.

To retrieve suitable search keywords, in addition to NLP-centric approaches like dependency parsing and TF-IDF, it is necessary to construct design knowledge graphs for specific streams such as engineering, architecture, software etc. Such knowledge graphs are likely to recommend new terms as well as assist with text completion for queries. For example, if we begin to search for 'bearing' and next-word predictions are 'lubricant', 'load' etc., we could choose 'bearing lubricant' and leverage from next word predictions like 'density', 'film', 'material' etc. Common-sense knowledge graphs like Google (and YouTube) make predictions based on many senses of the word 'bearing' and do not return the words as we have indicated in the example.

WordNet and ConceptNet have been the main supporting pillars for concept search as well as retrieval, while generic ontologies such as FBS and SAPPhIRE have been utilised to largely channel the search and retrieval processes. Since creative concepts emerge from the marriage of diverse sets of domains, a common-sense lexicon like WordNet is still a preferable supporting material. Similarly, scholars can also use readily available search methods like Google APIs to retrieve results from sources such as YouTube and patent databases. However, while retrieving concepts from a domain-specific knowledge source, it makes sense to utilise the domain-specific ontologies for query formation.

Alongside ontologies, scholars could benefit from the embeddings of common-sense language models like BERT and GPT-x[27] to obtain nearby search keywords, compare search results etc. Since the concept search and retrieval are largely exploratory and preferably involve diverse domains, the usage of common-sense language models shall not limit the desired performance of the NLP applications. While the same applies to concept association as well, the scholars shall also utilise domain-specific language models if the design problem is quite domain-specific.

Concept Selection involves one or more metrics like novelty, value, feasibility etc. Scholars could benefit from an affective lexicon to rate the design concepts and carry out systematic approaches to analyse and present the results. Since the theory behind these metrics is yet to be consolidated, the NLP applications are still in nascent stages. Scholars can only benefit from preliminary NLP tasks like similarity measurement, frequency analysis, and term retrieval to assist them with one or more steps in the concept selection process.

Design theorists could benefit from the NLP methods to examine how successful concepts are selected. For example, Arts et al. [116] observe the causality between frequencies of unigrams, bigrams, and trigrams and the likelihood of a patent getting an award, e.g., Nobel, Lasker, Bower, A.M. Turing etc. Similarly, scholars could leverage the text descriptions of concepts that have been selected for awards like Red Dot (indicates novelty or surprise), and Malcolm Baldrige National Quality Award (indicates value). Moreover, to understand the actual value of a concept, scholars could also utilise the sales information. For instance, Argente et al. [117] connect patent value with the number of product units sold from Nielsen's Retail Scanner data with the 'value' of a patent.

---

[27] https://beta.openai.com/

## 3.3. Discourse Transcripts

Design communication is often documented as discourse transcripts in protocol studies, think-aloud experiments, and recorded design workshops. Starting with a design issue or a problem, designers communicate sub-issues, solutions, related artefacts, arguments, justifications etc. Identifying and analysing the set of concepts that arise during such communications allows scholars to reveal a variety of insights about the design process.

A sequence of closely-related concepts within a segment in discourse transcripts represents a period of coherent communication, which could affect the design outcomes in terms of success metrics like novelty, feasibility etc [118]. To identify such concepts, scholars have extracted nouns, phrases, segments, topics etc., and associated these using vector-based or corpus-based similarity measurement techniques. We review such NLP-based approaches that are currently preferable over traditional linkographs [119].

*3.3.1. Concept Identification*

Scholars have adopted different approaches to extract key concepts such as topics, words, ontology-based entities, and n-grams. Wasiak et al. [120, p. 58] analyse emails to discover topics such as functions, performance, features, operating environment, materials, manufacturing, cost, and ergonomics. From the email exchanges in a traffic wave project, Lan et al. [121, p. 7] map word-frequency vectors and topic vectors (tasks, timestamps, persons, organizations, locations, input/output, techniques/tools) using Deep Belief Network – DBN [122], [123].

Goepp et al. [124, p. 165] identify the following speech acts from email exchanges: Information, Explication, Evaluation, Description, and Request. These speech acts were associated with a set of verbs, e.g., "Explication" was associated with 'explain', 'clarify' etc. To capture significant phrases that denote design changes, using the DTRS7 dataset, Ungureanu and Hartmann [125] extract n-grams (0 < n < 8) based on frequency analysis and examine how short terms progress to a variety of long terms; e.g., "a little" → "a little bit bigger", "a little splash of colour" [125, p. 12].

*3.3.2. Design Process Characterization*

To characterize the design process, scholars have aggregated the concepts thus identified from discourse transcripts into a whole (e.g., a semantic network) and perform analyses as reviewed below.

To characterize coherence in design communication, Dong [126, pp. 450, 451] obtain vector representations of emails and memos using LSA and measure the standard deviations of these w.r.t., their centroid (mean). A low standard deviation of the set of vector representations is considered to denote a high coherence in communication [127, p. 381]. In an alternative approach, Dong [128, pp. 39, 40] identifies segments by linking noun sequences using lexical relationships obtained from WordNet.

Based on the word occurrences of design alternatives within a time interval, Ji et al. [129] model the relationship between preferences using the Preference Transition Model and Utterance-Preference Model. Menning et al. [130, pp. 139, 142] use cosine similarity between LSA vectors of consecutive discourse entities to measure coherence. To characterize the uncertainty of the design process, Kan and Gero [131] measure the text entropy of the transcripts that were obtained from protocol studies.

Georgiev and Georgiev [132] utilise 49 WordNet-based semantic similarity measures to build a noun-based semantic network of students' and instructors' conversations as given in the DTRS10 dataset. They plot the average semantic similarity, information content, polysemy, and level of abstraction w.r.t., time for characterizing the design communication. Casakin and Georgiev [133] train regression models to establish a relationship between these network properties and the following metrics of design outcomes: originality, feasibility, usability, creativity, and value.

*3.3.3. Summary*

As we summarize in Table 4, the NLP applications built using discourse transcripts are limited in comparison with other types of text sources due to the least accessibility and information content. While emails do not strictly qualify as 'transcriptions' of design communication, the currently available data sources are mainly DTRS datasets. Scholars could additionally explore panel discussions, protocol studies and client interactions (e.g., architect and customer). The accessibility to such sources is crucial for the development of NLP applications regarding discourse transcripts.

Table 4: Summary of NLP methodologies and future possibilities with Discourse Transcripts.

|  | Data | Methods | Supporting Material |
|---|---|---|---|
| **Concept Identification** | Emails, DTRS7 Dataset, Protocol Studies | **Classification**: Deep Belief Network<br>**Embedding**: TF-IDF, LDA, BERT, GPT-x<br>**Segmentation**: N-grams, LDA<br>**Others**: Frequency Analysis | WordNet, Domain-specific ontologies |
| **Design Process Characterization** | Emails, Memos, DTRS10 Dataset, Protocol Studies, Panel Discussions, Client Interactions | **Similarity Measurement**: WordNet-based Similarity, Lexical Relationships, Preference Transition Model, Utterance-Preference Model, Cosine Similarity<br>**Embedding**: LSA<br>**Others**: Network Analysis, Linear Regression, Text Entropy | WordNet, Domain-specific ontologies |

Beyond the likelihood of obtaining one or more of these sources, the probability of extracting meaningful design knowledge from these is quite limited. For instance, the usage of frequent colloquial phrases like "sort of too big" limits the possibility of applying NLP methods to these [134]. Moreover, a transcription, unlike any text document, involves a timestamp associated with its parts. Several factors such as lack of context, poor grammar, colloquial language, time variation etc., are beyond what state-of-the-art NLP could handle. Scholars could still conduct preliminary analyses as they have done so far in terms of segment identification and topic discovery. Such analyses could also be benefitted from common-sense language models because verbal communication involves many common-sense terms. If required, scholars may also utilise domain-specific ontologies to recognize the domain terms in their analyses.

## 3.4. Technical Publications

Technical publications include over 92 million patents and a portion of over 174 million records that comprise textbooks, journal articles, and conference proceedings[28]. Due to their coverage, size, and accessibility, these sources carry a significant advantage over other sources in terms of knowledge aids for the design process. Moreover, since these sources are peer-reviewed and adhere to grammar and typographical norms, NLP tasks are well-suited to these [135].

*3.4.1. Patent Documents*

As mentioned in Section 3.1.3, design knowledge that is extracted from the text could be in the form of terms, phrases, segments etc., that should represent one or more constituents of an artefact that is relevant to the design process. Moreover, if re-represented, such terms, phrases, and segments of text must assume the form <entity, relationship, entity> to store these in a machine-readable form. Being a large body of technical inventions, patents offer a **rich source of design knowledge** that is also characterised by high information content, quality, and technicality.

---

[28] https://clarivate.libguides.com/webofscienceplatform/coverage

To extract design knowledge from patents, scholars have primarily utilised ontologies to channel their extraction approaches. To extract issue-related concepts and relationships (noun-noun, noun-adjectives), using a WordNet-based similarity, Liu et al. [40, pp. 4, 5] compare sentences in 46 patent abstracts against an ontology (list of terms) of issues, disadvantages, and challenges. Moehrle and Gerken [136] use a domain ontology to extract bigrams and trigrams from 522 patents of SUBARU's four-wheel drive. They use the terms thus extracted to measure patent-patent similarity using a variety of measures [136, p. 817] such as Jaccard, Inclusion, Cosine, DSS-Jaccard etc.

Liang et al. [137] adopt a sentence graph approach and Issue-Solution-Artefact ontology to extract design rationale from 18,920 Inkjet Printer patents that were assigned to Hewlett-Packard (HP) and Epson. Using a similar dataset of inkjet printer patents, Liang et al. [8] develop the Topic-Sensitive Content Extraction (TSCE) model and verify the model by testing the effect of segment length, parameters, sample count, and topic count etc. Fantoni et al. [77] propose a heuristic approach to extract the terms that correspond to functions, behaviours, and structures (FBS ontology) from patents. In *function*, for example, they consider the frequent combinations of verb-noun and verb-object.

To discover the structural form assumed by a collection of patents, Fu et al. [138] perform LSA on 100 randomly selected US patent documents. They consider only verbs (functions) and only nouns (surfaces) to perform two different LSAs and thereby obtain corresponding patent vectors. Using the cosine similarities, they discover the most optimal structural form – hierarchy using which they construct a patent network. They also label the clusters of patents with the closest terms (verbs or nouns).

Upon training the abstracts of 500,000 patents (CPC-F subsection) using Word2Vec, Hao et al. [139] obtain the embeddings of 1700 function terms (e.g., grill, cascade) that are given by Murphy et al. [140]. They obtain a patent vector as a circular convolution ($\otimes$) of function terms that are present in the corresponding patent abstract. To support efficient retrieval of patent images, Atherton et al. [141, pp. 247, 248] annotate images in USPTO with geometric features and functional interactions extracted from claims. Song and Fu [142] obtain three patent-word matrices using 1,060 patents and three sets of words corresponding to components, behaviours, and materials. They apply a Non-negative Matrix Factorization algorithm to these matrices to extract significant topics.

While patents offer design knowledge in specific domains, due to the totality of technology space covered by the patent database, scholars have also attempted to construct WordNet-equivalent lexicon as well as engineering ontologies. Vandevenne et al. [143, p. 86] analyse titles and abstracts in a randomly drawn set of 155,000 patents from the EPO database[29] to discover that nouns are abstract (e.g., system, device) and are meaning enablers (e.g., temperature, pressure) that also point towards the product (e.g., valve, display).

To identify the primary users of technological inventions that are documented as patents, Chiarello et al. [144] extract a generic list of users in terms of job positions, sports, hobbies, animals, patients, and others. They identify these generic users in selected patents[30] using a semi-automatic approach and annotate the sentences using these. They then feed the annotated dataset of sentences into SVM and Multi-Layer Perceptron for Named Entity Recognition (NER).

Sarica et al. [79] obtain embeddings of over 4 million unique terms from the titles and abstracts across the US patent database. Using a web-based tool called TechNet[31], they facilitate a search for these terms [145] and utilise the embeddings of these to construct a similarity network [146]. To create an engineering alternative to WordNet, Jang et al [78] collect 34,823 automotive patents (IPC B60). They examine the dependency patterns in abstracts and claims to extract dependency relations that form the TechWord network. For the words in the network, they create TechSynset by capturing the WordNet synonyms and calculating the cosine similarity between BERT-based embeddings of individual pairs.

Scholars have demonstrated how patents could act as **stimuli for generating concepts** as well as indirect supports for problem-solving through TRIZ-based tools [147]–[150]. In the effort to discover patent network structures, Fu et al. [138] include a design problem in their LSAs to identify a starting point for navigating the patent network. Given a starting point in the network, Fu et al. [151] consider patents at one and three hops as 'near' and 'far' respectively. They examine the effect of 'near' and 'far' patents on novelty and quality when these patents are given stimuli alongside the design problem.

---

[29] https://www.epo.org/searching-for-patents.html

[30] The selected patents fall within the A47G33, A61G1-G13 groups that are defined by the International Patent Classification (IPC).

[31] http://www.tech-net.org/

To support patent retrieval, Murphy et al. [140] adopt a Zipfian statistic approach to extract 1700 function (verbs) terms from 65,000 patent documents and organise these into primary, secondary, and tertiary w.r.t., the functional basis [152]. They index 2,75,000 patents using these functions that also act as query elements. To map design problems and patents via the Functional Basis, Longfan et al. [153] train a semi-supervised learning algorithm based on Naïve Bayes and E-M algorithm using 1666 patents and the texts labelled with function categories. In another approach, they extract meaningful terms from patents using a frequency-based statistic [153, p. 8] and cluster the patents according to the terms.

Although several approaches to searching and managing patents exist [154]–[156], it is necessary to simplify the patent documents before utilising these as stimuli for generating concepts. To form keyword summaries of patent search results, Noh et al. [11] conduct an experimental study to find that it is best to extract 130 keywords from abstracts using TF–IDF and Boolean expression strategies.

Sarica et al [145] propose TechNet [79] as a means to search and expand technical terms, which were extracted from the titles and abstracts in the patent database. To facilitate cross-domain term retrieval, Luo et al. [157] organise the output of TechNet into various domains that are associated by a knowledge distance measure. Souza et al. [13] train an LSTM-based sequence-to-sequence (abstract-title → summary) neural network using 7000 patents for generating abstract summaries of patent documents. They group the summaries thus generated using a semantic similarity measure [158] and subsequently identify patent clusters.

Patents not only document technological inventions but are also assigned to specific domains, companies, inventors, countries etc. Using such meta-data, scholars have developed **technology maps for exploring design opportunities**. Jin et al. [159] extract meaningful terms from patents and use Bag-Of-Words (BOW) approach to create patent vectors that form a technology map. Trappey et al. [160] adopt a similar approach to patents and clinical reports that concern dental implants. Altuntas et al. [161] use the same dental implant patents and obtain vectors of these using the patent-class matrix. They cluster the patent vectors using the following methods: E-M algorithm, Self-organizing map, and Density-based method.

To explore new design opportunities as well as to aid in idea generation, Luo et al. [162] develop a technology space map using all CPC 3-digit classes and the co-citation proximity measures among these. They implement the map using support called InnoGPS[32] which provides several interactive features that are analogous to Google Maps. The support tool mainly allows the users to position themselves on the technology map, identify the closest domains, and navigate the technology space map. Luo et al. [163] conduct an experimental study to demonstrate how the total technology space map is useful for exploring "white space" design opportunities related to Artificial Neural Networks and Spherical Rolling Robots.

To identify new technology opportunities relating to carbon-fibre heating fabric, Russo et al. [164] download 16,743 patents and extract Subject-Action-Object triples where the Subject is "heating fibre". Assuming that Action represents a function, they mine dependency patterns to extract applications (e.g., 'applied as', 'used for') and requirements (e.g., 'enhance…', 'un…ability') pertinent to the heating fibre technology. To explore new technology opportunities using products, Lee et al. [165] use the patent-product database[33] developed at the Korea Institute of Science and Technology Information (KISTI). They extract Word2Vec embeddings for products and technologies to create an exploration map that allows the identification of technologies closer to products and vice-versa. They also propose 10 indices to assess the performance of technology exploration.

To identify technology opportunities in 3G that could be leveraged in 4G, Zhang and Yu [166] extract effect phrases from the corresponding patents using a Bi-LSTM with a conditional random field layer. They label the words in the sentences using {Begin, Inside, Other} of an effect phrase and feed the labelled data into the neural network. They combine the effect phrases in a patent using a weighted TF-IDF vector and use topic clustering to group the patents. Depending on the number of patents on each topic, they calculate the technology opportunity score [166, pp. 560, 561].

*3.4.2. Textbooks and Handbooks*

Several design studies support the notion that exploring concepts from distant domains could lead to novel design solutions. Adhering to this consensus, Shu and colleagues have conducted analyses on a biological textbook [167] to understand the characteristics that support bio-inspiration. Shu [17, p. 510] understands that the textbook includes candidates for design-by-analogy, e.g., ('bacteria', 'fill', 'pores of clothes') →

---

[32] http://www.innogps.com/
[33] https://repository.kisti.re.kr/handle/10580/14535

"prevent dirt". Cheong et al. [82, pp. 4, 5] identify that in the text, domain and common verbs co-occur, e.g., "*received* and converted or *transduced*".

To capture causally related biological functions, Cheong and Shu [168, pp. 1–4] locate and extract pairs of enabler-enabled functions using syntactic rules, e.g., "Lysozymes *destroy* bacteria to *protect* animals". Upon searching in the same textbook, Lee et al. [86, pp. 5–7] identify morphological nouns that co-occur with the keywords in a single paragraph. For every noun, they calculate a modified TF-IDF metric [86, pp. 5, 6] for usage in LSA.

The following articles describe approaches to extracting design knowledge from published handbooks. Hsieh et al. [169] mine the Table of Contents, Definitions, and Index from an Earthquake Engineering Handbook to develop a domain ontology. Kestel et al. [170] apply several text mining steps to the published document that describes the standard procedure for simulation of multi-bolted joints (VDI 2230 Part 2). They extract structured data with specific attributes (e.g., part, contact, load, relation) from the text and utilise these to build ontologies that are integrated with Finite Element Analysis (FEA) tools.

Richter et al. [171] obtain the design standards and guidelines for landfilling in different provinces of Canada. They conduct word frequency analyses using metrics such as Gunning-Fox Index and Lexical Density. Xu and Cai [80] mine 300 sentences from the underground utility accommodation policies from the departments of transportation such as Indiana and Georgia. They use utility-product and spatial ontologies to process and label the terms in the sentences with seven categories [80, p. 7]. They examine the POS and category patterns in these sentences to extract hierarchical knowledge structures.

*3.4.3. Scientific Articles*

Unlike patents and books, the overall motive behind processing scientific articles is unclear, mainly due to a limited number of contributions. We therefore review these contributions as follows by explicitly stating the purpose beforehand. To summarize engineering articles by discovering their micro-and macro-structures, Zhan et al. [172, pp. 5, 6] train Naïve Bayes and SVM classifiers by labelling 1425 sentences from 246 research articles into one of the four categories: background, contribution, methodology, results and conclusions.

To identify the sentences that could aid in bio-inspiration, Glier et al. [173, pp. 5–7] represent sentences from five biology journals using a feature vector of 1,869 terms and label these as 'useful' and 'not useful' for bio-inspiration. They feed the labelled dataset into the following classifiers: SVM, k-NN, and Naïve Bayes. To build a bridge between biological and engineering domains and thus aid bio-inspiration, Vandevenne et al. [143, p. 82] map product and organism aspects upon processing 155,000 EPO patents and 8,011 papers from the Journal of Experimental Biology.

To create a generic engineering ontology, Shi et al. [174, pp. 4–6] develop a large semantic network called B-Link by extracting and combining entities from technical websites and articles, respectively, using Scrapy[34] and Elsevier APIKey[35]. To understand the evolution of typology in design research, education, and practice, Won and Park [175] collect 222 terms[36] from over 300 documents that include design publications, abstracts etc. and discover that these terms have evolved from being object-based to concept-based.

To understand the definitions of contemporary technologies such as Artificial Intelligence, Industry 4.0 etc., Giordano et al. [176] identify these terms in the sentences of Elsevier-Scopus abstracts and filter the cases where the neighbour of these terms adhere to a pattern, e.g., "defined as", "refer to" [176, p. 10]. They further analyse the frequencies of the constituents of these sentences so filtered. To understand the field of Product-Service Systems (PSS), Rosa et al. [177] develop a concept map by analysing 29 articles relating to the design of PSS.

*3.4.4. Summary*

We have summarized the NLP contributions that use technical publications in Table 5. Due to high accessibility, information content, quality, and technical density, technical publications have been quite popular sources for developing NLP applications. The methodologies have also adopted state-of-the-art NLP methods while also utilising domain ontologies wherever applicable. Therefore, a little could be commented about the potential gaps in these contributions.

---

[34] https://scrapy.org/

[35] https://www.elsevier.com/solutions/sciencedirect/support/api

[36] https://www.tandfonline.com/doi/suppl/10.1080/14606925.2021.1906085/suppl_file/rfdj_a_1906085_sm7942.docx

Table 5: Summary of NLP methodologies and future possibilities with Technical Publications.

| | Data | Methods | Supporting Material |
|---|---|---|---|
| **Patent Documents - Design Knowledge Extraction** | **Patents**: SUBARU 4-Wheel Drive, HP Inkjet Printers, Epson Inkjet Printers, IPC-A47G33, IPC-A61G1-A61G13, IPC-B60<br>**Others**: Career Planner, Not-so-Boring Life, Discover a Hobby, A-Z-Animals, Medicine Net, Centre for Disease Control and Prevention | **Similarity Measurement**: Latent Semantic Analysis, WordNet-based Similarity, Patent-Patent Similarity (e.g., Jaccard), Dependency Parsing, BERT, Domain-Specific Language Model<br>**Term Retrieval**: Topic Sensitive Content Extraction, Rule-based Mining, Dependency Parsing, WordNet synonyms<br>**Relation Extraction**: Rule-based Approach<br>**Topic Extraction**: Non-Negative Matrix Factorization<br>**Named Entity Recognition**: SVM, Multi-Layer Perceptron<br>**Others**: Bayesian Networks | **Lexicon**: WordNet, TechNet, TechWord<br>**Ontology**: Issues, 4-wheel drive, Issue-Solution-Artefact, FBS |
| **Patent Documents - Concept Generation Stimuli** | **Patents**: CPC-F | **Embedding**: LSA, Word2Vec, Circular Convolution, Domain-Specific Language Model<br>**Similarity Measurement**: Bayesian Network, Semantic Similarity<br>**Classification**: Naïve Bayes, Expectation-Maximization (E-M)<br>**Text Generation**: LSTM | **Lexicon**: TechNet, TechWord<br>**Ontologies**: Functional Basis |
| **Patent Documents - Design Opportunity Identification** | **Patents**: Dental Implants, KISTI Patent-Product Database, 3G, 4G<br>**Others**: Clinical Reports | **Embedding**: Bag of Words (BOW), Patent-Class Matrix, Word2Vec, TF-IDF<br>**Clustering**: E-M Algorithm, Self-organizing Map, Density-based Approach<br>**Classification**: Bi-LSTM CRL | **Ontologies**: International Patent Classification |
| **Textbooks and Handbooks** | Biology Textbook, Earthquake Engineering Handbook, VDI 2230 Part 2, Landfilling Guidelines, Underground Utility Accommodation Policies | **Term Retrieval**: Rule-based Approach, Gunning-Fox Index, Lexical Density, POS patterns, Category Patterns<br>**Embedding**: TF-IDF, LSA | **Lexicon**: WordNet<br>**Ontologies**: Underground Utilities |
| **Scientific Articles** | **Articles**: Computer Integrated Manufacturing, Basic and Applied Ecology, Current Biology, Journal of Animal Behaviour, Journal of Animal Ecology, Journal of Zoology, Journal of Experimental Biology, ScienceDirect, Design Publications, Design of Product-Service Systems (PSSs), Conference Proceedings<br>**Patents**: European Patent Office | **Classification**: Naïve Bayes, SVM, k-NN<br>**Term Retrieval**: Rule-based Approach<br>**Topic Extraction**: LDA variant<br>**Relation Extraction**: Supervised Approach | |

Scholars could invest more effort into scientific articles (including conference proceedings) as the literature on patent analyses is extant. In addition, scholars could also report more analyses on full texts of patent documents, as a majority of contributions are limited to titles, abstracts, and claims. Scholars could leverage the wealth of knowledge in these sources to create ontologies and knowledge graphs both at the generic and domain-specific levels. As a part of knowledge graph extraction, relation extraction shall adopt a rule-based approach in patent documents as the language is consistent across the entire database. In scientific articles, however, relation extraction requires prior named entity recognition as well as relation label prediction algorithms. Scholars could also immix patent documents and scientific articles in a particular domain to develop a domain-specific graph extraction tool.

## 3.5. Consumer Opinions

Available in plenty as a part of social media text and product reviews, consumer opinions are reflective of actual user experiences [178], product specifications, requirements, and issues [179]. Consumer opinions often include typographical errors (e.g., coooolll), alternative word forms (e.g., LOL), multi-lingual terms, and grammatical errors. It is a challenge to remove symbols, hyperlinks, usernames, tags, artificially generated messages, and misspelt words. Lim and Tucker [180, pp. 1, 2] posit that identifying product features in consumer opinions often involves challenges in term disambiguation (e.g., "researchers should really screen for this type of error") and keyword recognition ("…just as this court case is about to start, my iPhone battery is dying").

To work around the above-mentioned issues, Tuarob and Tucker [181] propose using Carnegie Mellon POS tagger that suits social media text. In addition, He et al. [106, p. 4] recommend using TextRazor[37] for identifying proper nouns like 'Uber' and 'Manhattan'. While processing consumer opinions, Tuarob et al. [182, p. 4] prefer not to perform stemming due to its negative effects on the performances of downstream NLP tasks. To improve the grammatical structure, Wang et al. [183, pp. 456–458] suggest a few transformation rules, e.g., Sentence 1 (e.g., "very nice") is prepended with subject and verb to obtain Sentence 2 (e.g., "It is very nice") if the former does not include these. In addition to these approaches to work around the issues with consumer opinions, scholars have incorporated distinct steps before performing sentiment analysis, capturing usage context, and modelling user emotions.

*3.5.1. Sentiment Analysis*

Sentiment analysis is an important application of NLP that uses ratings as well as an affective lexicon to determine the polarity and intensity of sentiment in a piece of text. The sentiment scores quantify the product favourability [181, p. 5] and affective performances [184, pp. 450, 451]. Obtaining true sentiment scores is often a challenge, given that 22.75% of a social media text is sarcastic [182]. In addition, Tuarob and Tucker [185] identify that neutral words constitute over 53% and 48.6% of smartphone and automobile-related tweets. Since the sentiment score of a phrase may not often match that of a sentence, Chang and Lee [184,

---

[37] https://www.textrazor.com/

p. 462] propose to adjust the sentiment score of a local context based on the polarity match with the whole sentence.

Sentiment analysis utilises product features (nouns) and sentiment indicators (adjectives, adverbs, and verbs); e.g., "The keyboard is fine but the keys are real slippery" includes product features {keyboard, keys} and sentiment indicators {fine, slippery} [186, pp. 1, 2]. Sentiment analysis requires an affective lexicon like SentiWordnet [187], Affective Space 2 [188], and SenticNet 6 [189]. We review in the remainder of this section, the contributions that have conducted sentiment analyses on various design text sources.

Ragupathi et al. [190] compute sentiment scores of Home Theatre reviews from Twitter, Amazon, and Flipkart using the SENTRAL algorithm and the Dictionary of Affect Language – DAL. To predict sentiment scores, Zhou et al. [191, p. 4] feed a labelled dataset of Kindle Fire HD 7 reviews into the fuzzy-SVM algorithm along with a lexicon that is populated using ANEW [192]. Jiang et al. [193, pp. 2, 4] extract nouns, adverbs, verbs, and adjectives from electric iron reviews and utilize SentiWordNet [187] to predict sentiment scores.

Zhou et al. [194] compute sentiment scores of specific product features in Kindle Fire HD reviews using ANEW and classification based on a rough set. They augment the sentiment scores with a feature model that was constructed by extracting product features using ARM and combining these using WordNet-based similarity measures (e.g., Resnik, Leacock-Chodorow). Jiang et al [195, p. 394] assess 1259 reviews of six compact cars using Semantria[38] to obtain sentiment scores. Tuarob et al. [182, pp. 6, 8] use TextBlob[39] to compute sentiment scores of tweets related to 27 smartphone models. They account for sarcasm using the analysis of a coword network, where nodes are ranked for likelihood, explicitness, and relatedness.

Tang et al. [186] develop the Tag Sentiment Aspect (TSA) Model to extract topics and sentiment indicators simultaneously. They demonstrate the proposed TSA model using DSLR and Laptop reviews [196]. Sun et al. [197] calculate sentiment scores of 500,000 phone reviews from Zol[40] upon capturing the co-occurrence of product features and sentiment indicators (adjectives, adverbs) within a sliding window. For sentiment analysis, Suryadi and Kim [198, pp. 3, 4] feed the labelled embeddings of informative Laptop and Tablet Amazon-based reviews into the Long-Short Term Memory (LSTM) model [199].

---

[38] https://www.lexalytics.com/support/apps/excel
[39] https://textblob.readthedocs.io/en/dev/
[40] https://www.zol.com.cn/

Sun et al. [200] mine 98,700 reviews and product descriptions of Trumpchi GS4 and GS8 vehicles that are manufactured by the GAC group. They use TF-IDF and fastText [201] to compute sentiment scores and extract attributes from the text thus mined. Chiarello et al. [202] extract 7,165,216 Twitter posts that appeared ahead of the launch of Xbox One X and New Nintendo 2DS XL to examine the effect of sentiment polarity of the social media activity upon the success of these products. They label 6,500 tweets relevant/irrelevant and build an SVM classifier. Upon classifying the tweets that are outside the training set, they obtain 66,796 relevant tweets and compute the sentiment scores of these using a specific lexicon [203].

Gozuacik et al. [204] classify Google Glass tweets using a Deep Neural Network for sentiment polarity and opinion usefulness. They include bag-of-words and other embedding techniques for comparing the classification performances. They find using clustering analysis [204, pp. 9–11] that among the useful opinions, negative ones denote issues and positive/neutral ones denote innovations. To identify sentiment indicators, Han and Moghaddam [205] collect 23,564 sneaker reviews and fine-tune BERT for a Named Entity Recognition task with the following labels on each word in a sequence: background, sentiment, attribute, and description.

Han and Moghaddam [206] extract product attributes of sneakers from catalogues and product descriptions and apply a rule-based approach to compute sentiment scores w.r.t., these attributes. Li et al. [207] identify groups of customers and attribute preferences by clustering the Word2Vec embeddings of 30,000 laptop reviews from JD[41]. They estimate the sentiment score using Microsoft's Deep Structured Semantic Model and utilise these sentiment scores to develop a Kano map between customer groups and attribute preferences.

*3.5.2. Extracting Usage Context*

In this section, we review the contributions that capture usage context by examining the product features and their functioning in different environments [208]–[210]. Park and Lee [211] extract consumer opinions of 135 mobile phone models from a review portal[42]. Upon analysing the opinion data using TextAnalyser 2.0, they mine the frequent product specifications, cluster the consumers, and form product-specific networks.

---

[41] https://global.jd.com/

[42] http://www.mobilephonesurvey.com

Wang et al. [212] label and group camera reviews from Amazon and NewEgg using the frequent keywords obtained from product descriptions. They extract the aspects from these reviews using Fine-Grained LDA and Unified Fine-Grained LDA. To relate engineering characteristics with consumer opinions, Jin et al. [213] obtain 770 reviews of HP and Epson printers from Amazon to extract engineering characteristics using n-gram language models. To aid House of Quality construction, Ko [214] relate consumer and design requirements using a 2-tuple fuzzy linguistic approach.

To extract important product features, Jin et al. [215] select the most representative sentences from 21,952 reviews on CNET using a greedy algorithm and verify these using information comparativeness, information representativeness, and information diversity. To classify product reviews, Maalej et al. [216] procure over 1.2 million Smartphone Application reviews from the Apple AppStore and Google Play Store. They label the reviews according to four categories: bug report, feature request, user experience, or rating and train the labelled dataset using Naïve Bayes, Decision Tree, and Maximum Entropy algorithms, while also examining the effect of different approaches such as Bag of Words, Bigrams, Lemmatization, Stop words etc.

To extract product usage, Park et al. [217, p. 4] learn feature ontology by measuring triples like "fabric + shrink" using Wu and Palmer similarity [218] and merging with factual (e.g., "fabric + rayon") and sentiment (e.g., "shirt + disappoint") ontologies using a Fuzzy Formal Concept Analysis (FFCA) approach. They identify the relationships between triples using explicit causal conjunctions like 'so', 'due to', 'because' etc [217, p. 6].

To disambiguate product reviews, Singh and Tucker [219] classify the Amazon review sentences (obtained using import.io) into function, form, behaviour, service, and other using the following classifiers: Naïve Bayes, SVM, Decision Tree, and IBk classifiers. To identify the type of design knowledge in a product review, Kurtanovic and Maalej [220] label 32,414 reviews of 52 Amazon Store Apps with the following concepts: Issue, Alternative, Criteria, Decision, and Justification. They apply the labelled dataset to the following classification algorithms: Naïve Bayes, SVM, Logistic Regression, Decision Tree, Gaussian Process, Random Forest, and Multilayer Perceptron.

To capture bigrams that represent the usage context of wearable technology products, Suryadi and Kim [221, pp. 6, 7] combine noun-adjective pairs that co-occur in a hierarchical path in the dependency tree. They [221, p. 8] group the embeddings of the noun-adjective combinations using $X$-means clustering. In an extended work, Suryadi and Kim [198, p. 7] identify bigrams that are noun-verb, noun-noun, while verbs end with a -ing; e.g., 'web browsing', 'reading books'.

Hou et al. [208, p. 3] structure an affordance description as "Afford the ability to [action word] [action receiver] [perceived quality] [usage context]". Based on the structure, they [208, p. 5] extract perceived opposite qualities (e.g., low, high) from Kindle reviews to train an ordered logit regression model. An affordance $i$ that supposedly has the perceived qualities $X_i$ and $Y_i$ is characterized according to their model by the coefficients $\alpha_i$ $and$ $\beta_i$ that are used to identify categories of Kano [222] model: must be, performance, attractive, indifferent, reverse, questionable.

Zhou et al. [223] filter uninformative reviews of Amazon products like Echo, Alexa etc., using a fastText classifier and extract topics from these using LDA. To estimate the importance of product attributes, Joung and Kim [224] collect 33,779 smartphone reviews from Amazon. They identify product attributes using LDA and sentiment scores using IBM Watson. They estimate the importance of product features using k-optimal Deep Neural Networks that were designed using the SHAP[43] method.

*3.5.3. Kansei Engineering*

Kansei engineering aims to support the emotion-driven design and involves acquisition of emotional responses using bipolar adjectives like 'hot-cold' and 'unique-conventional' [113], extracting descriptive adjectives like 'fresh' and 'appealing' [225], and clustering these adjectives [226], [227]. The NLP contributions as we review in the remainder of this section involve developing emotion vocabulary, describing emotions of artefacts, modelling product features and emotions, and developing fuzzy-linguistic membership functions.

Scholars have proposed **design-specific emotion vocabulary** to characterize artefacts. Desmet [228, pp. 4, 5] proposes 9 groups of 25 emotion types and representative emotion words within these. Chaklader and Parkinson [229, pp. 2–4] examine 500 reviews of Bose SoundLink headphones to identify 29 cue terms [229, p. 2] that reflect ergonomic comfort. Kim et al. [230] identify 15 clusters of 4,941 reviews of recliners from Amazon and extract the most frequent adjectives from these clusters.

Scholars have applied **existing vocabulary to describe artefacts** in their studies that we review as follows. Karlsson et al. [231] use several adjectives to describe the interiors of BMW 318, Volvo S60, VW Bora and Audi A6 along the lines of the following factors: pleasantness, complexity, unity, enclosed-ness, potency, social status, affection, and originality. To identify the extent of brand importance in the design process, Rasoulifar et al. [232, pp. 144, 145] interview 30 designers about a Tecnifibre tennis bag. From the responses, they extract Kansei, design and brand concepts and organise these into a multiple domain matrix.

---

[43] SHapley Additive Explanations - https://shap.readthedocs.io/en/latest/index.html

To characterize the affective qualities of electronic readers, Wodehouse et al. [233, pp. 489–492] obtain descriptive adjectives of these from a survey on visual attractiveness. They use RAKE[44] to extract keyphrases (e.g., "prefer physical books") from the patent documents relevant to electronic readers. They form feature vectors of electronic readers using descriptive adjectives and the key phrases to cluster these vectors using ClusterGrammer[45].

To compare affective performances of similar products, Liao [234, p. 5/18] ask survey participants to place eight wearable products on the quadrants of two graphs: comfortable vs. like clothing and delightful vs. like clothing. Upon placing the products, they also ask the participants to select a suitable emotional descriptor [234, p. 8/18]. Hu et al. [235] collect emotional responses of a flash drive regarding its colour, contour, and shell material to discover the emotional dimensions via multi-factor analysis. Using a case study on Toaster, Guo et al. [236] assess Kansei ratings of groups based on consensus and dominance

Scholars have attempted to establish a relationship between **emotional descriptors and product features**. Using a dataset of 7 interior designs of truck cabs, Zhou et al. [237] adopt K-optimal rule discovery and Ordinary Least-Squares Regression to map design elements and affective descriptors. Upon obtaining participant data on CNC machine tools, Wang [238] establish a relationship between abstract (e.g., "Rigid/Flexible") and elementary (e.g., "Firm/Fragile") Kansei words using Support Vector Regression and Artificial Neural Network.

Vieira et al. [239] measure the actuation force, contact force, stroke, and snap ratio for 11 keys in an in-vehicle rubber keypad. They ask participants to rate the performances of these keys using 7 Kansei words (e.g., unpleasant/pleasant, smooth/hard, loose/stiff). They observe using regression models that a significant relationship exists between the aforementioned design parameters and the Kansei ratings. To predict the Kansei ratings from the features of a bottle design, Mele and Campana [240] train a neural network with the following architecture: input layer with 14 design features (e.g., geometry, process, material), two hidden layers, and an output layer with eight ratings to corresponding Kansei words (e.g., classic/trendy, masculine/feminine).

---

[44] https://pypi.org/project/rake-nltk/
[45] https://clustergrammer.readthedocs.io/

Misaka and Aoyama [241] obtain Kansei ratings of crack patterns on pottery surfaces using 50 adjectives. They use a neural network with one hidden layer to model the relationship between ratings and crack characteristics such as width, fineness, and fluctuation. Upon mining 4459 Amazon reviews of 30 road bikes using WebHarvy[46], Chiu and Lin [242] construct a functional model and a morphological matrix for six design elements (e.g., saddle, tread surface). They identify the 11 most frequent adjectives and group these into four semantic sets (overall impression, usability, riding experience, and weight) and compute the corresponding semantic differentials. They run a linear regression using each semantic differential as a dependent variable and the six design elements as binary categorical variables.

So [243] conducts a study that involves ranking 115 adjectives to obtain 12 design words and five emotion words. Using the resultant words, he performs factor analysis to discover the following dimensions: tool, novelty, energy, simplicity, and emotion. Among these dimensions, he found that emotion was a significant predictor of design preferences via the following models: Linear Regression, Random Forest, Neural Network, and Gradient Boosting Machine. For 1474 French Press coffee maker reviews on Amazon, El Dehaibi et al. [244, pp. 4–6] use crowdsourced efforts to highlight phrases that indicate sustainability and obtain the corresponding degree of emotion. They train a logistic classifier to predict the DoE from highlighted phrases, while also using LDA to extract topics from these.

Wang et al. [245] propose rules to automatically label reviews with affective attributes (e.g., like-dislike, reliable-unreliable) based on the affective words contained in these. In an alternative approach to automatically labelling reviews, they build a classifier by manually labelling 900 reviews of 20 stuffed toys from Amazon and training the following models: k-NN, Classification and Regression Tree (CART), Multilayer Perceptron, DBN, and LSTM. Jiang et al. [246] extract hair dryer reviews from Amazon and estimate the predictability of product attributes (weight, heat, power, speed) upon minimum, maximum, and average sentiment scores over four time periods.

---

[46] https://www.webharvy.com/

Upon mining reviews and product specifications of 19 upper limb rehabilitation devices from Amazon and Alibaba, Shi and Peng [247] connect these with 10 customer requirements (e.g., flexible wear, no smell) using WordNet-based similarity. For each customer requirement, they measure satisfaction using adjectives and adverbs in the reviews. They also identify the functional implementation through product specifications. Next, they fit a curve to establish a relationship between functional implementation and customer satisfaction.

Chen et al. [248, pp. 84, 85] obtain 60 images of cockpit interior designs from the web and 20 emotional terms (about cockpit) from aircraft experts. They form the similarity matrix among these 20 terms using WordNet and cluster these into four emotional dimensions, which are used to rate each image as per the Likert scale [248, pp. 90, 91]. They train the following neural networks using the images labelled with an emotional degree: Radical Basis Function, Elman, and General Regression.

Kansei attributes [249, pp. 408, 409] and degrees [249, p. 410] are abstractions of adjectives (affective characteristic) and adverbs (affective degree). Using the affective degrees obtained from surveys or text mining, scholars have attempted to model the **linguistic membership functions** of affective characteristics. Wang et al. [249, p. 411] extract adjective-adverb combinations from McAuley's dataset[47] and map these to corresponding Kansei attributes and degrees. Wang et al. [250] map a variety of fuzzy linguistic term sets (e.g., {'none', 'very bad', 'bad', 'medium', 'good', 'very perfect', 'perfect'}) to their membership degrees using a trapezoidal asymmetric cloud model. Scholars have adopted similar approaches to model Kansei variables and their corresponding fuzzy membership functions, e.g., USB flash drives [251] and hand-painted Kutani cups [252].

*3.5.4. Summary*

We summarize the NLP contributions that use consumer opinions in Table 6. These sources have been quite popular alongside technical publications, given the extensive accessibility and high information content. However, consumer opinions are quite poor in terms of language quality, which, as discussed previously, poses negative effects on the performance of fundamental NLP tasks. Since prescriptive tools like NLTK do not work well on these sources, NLP scholars have been developing deep learning models to carry out fundamental tasks like POS tagging [253].

---

[47] http://jmcauley.ucsd.edu/data/amazon/

Table 6: Summary of NLP methodologies and future possibilities with Consumer Opinions.

|  | **Data** | **Methods** | **Supporting Materials** |
|---|---|---|---|
| **Sentiment Analysis** | **Amazon**: Home Theatre, Kindle Fire HD, Electric Iron, DSLR, Laptop, Hair Dryer, Tablet, Smartphone, Sneaker, Upper Limb Rehabilitation Device<br>**Twitter**: Home Theatre, Smartphone, Xbox One X, New Nintendo 2DS XL, Google Glass<br>**Others**: Flipkart – Home Theatre, Compact Car, Zol – Phone, Trumpchi GS4 & GS8, JD – Laptop, Alibaba – Upper Limb Rehabilitation Device, YouTube | **Text Processing**: Carnegie Mellon POS Tagger, Stanford CoreNLP, ARM<br>**Embedding**: TF-IDF, BOW, Word2Vec, GloVe, BERT<br>**Similarity Measurement**: WordNet-based Similarity<br>**Sentiment Prediction**: SENTRAL, Semantria, TextBlob, IBM Watson, Deep Structured Semantic Model<br>**Topic Extraction**: Tag Sentiment Aspect (TSA), LDA<br>**Clustering**: K-means<br>**Classification**: Fuzzy-SVM, LSTM, fastText, SVM, Deep Neural Network (DNN), Kano<br>**Named Entity Recognition**: BERT + 2 CNN layers<br>**Others**: Coword Network Analysis, Curve Fitting | **Lexicon**: WordNet<br>**Affective Lexicon**: SentiWordNet, WordNet-Affect, SenticNet4, Dictionary of Affect Language (DAL), ANEW |
| **Extracting Usage Context** | **Amazon**: Camera, HP Printers, Epson Printers, Smartphone, Kindle, Echo, Alexa<br>**Others**: MobilePhoneSurvey – Mobile Phone, NewEgg – Camera, CNET, Apple App Store, Google Play Store, YouTube | **Text Processing**: TextAnalyser 2.0, N-gram Model<br>**Term Retrieval**: Dependency Parsing, Rule-based Approach, Design Knowledge Graph<br>**Embedding**: BOW, Word2Vec, BERT, GPT-x<br>**Similarity Measurement**: Wu and Palmer<br>**Clustering**: K-means, X-means<br>**Topic Extraction**: LDA, Fine-Grained LDA, Unified Fine-Grained LDA<br>**Classification**: Naïve Bayes, Decision Tree, Maximum Entropy, SVM, IBk, Logistic Regression, Gaussian Process, Random Forest, Multi-Layer Perceptron, Ordered Logit Regression, Kano, fastText, k-optimal DNNs, SHAP<br>**Sentiment Analysis**: IBM Watson<br>**Others**: Fuzzy-Linguistic Approach | **Lexicon**: WordNet<br>**Ontologies**: Domain-specific |
| **Kansei Engineering** | **Amazon**: Bose SoundLink, Recliner, Road Bike, French Press Coffee Maker, Stuffed Toys, Hair Dryer, Upper Limb Rehabilitation Device (also from Alibaba)<br>**Survey Responses**: BMW 318 Interiors, Volvo S60 Interiors, VW Bora Interiors, Audi A6 Interiors, Tecnifibre Tennis Bag, Electronic Reader Patents, Wearable Products, Flash Drive, Toaster, Truck Cab Interiors, CNC Machine Tools, In-Vehicle Rubber Keypad, Bottle Design, Crack Patterns on Pottery Surfaces, Cockpit Interiors, Kutani Cups | **Clustering**: K-means, ClusterGrammer<br>**Classification**: K-optimal Rule Discovery, Ordinary Least Squares (OLS) Regression, Support Vector Regression, ANN, Neural Network (2 hidden layers), Linear Regression, Random Forest, Gradient Boosting Machine, k-NN, Classification and Regression Tree (CART), Multi-Layer Perceptron, DBN, LSTM, Radical Basis Function, Elman<br>**Topic Extraction**: LDA, RAKE<br>**Others**: Multi-Factor Analysis, Curve Fitting | **Lexicon**: WordNet |

Scholars have applied state-of-the-art NLP methods for sentiment analysis and extraction of usage context. Although Kansei engineering only concerns emotional descriptors for artefacts, scholars have significantly advanced this area by relating with product features and developing fuzzy linguistic models. While scholars could additionally explore the YouTube platform for a newer set of opinions, any advancement in NLP applications to consumer opinions, therefore, depends on the advancement in core NLP research.

The current NLP applications use state-of-the-art methods that can identify negative reviews, filter the less useful ones, extract significant topics, and group similar reviews. Companies can hire human resources to conduct *post hoc* analyses and test the products and services under those conditions that the consumers had deemed to malfunction. Developing NLP applications to support such *post hoc* analyses may not carry scholarly merit as much as generating value for the industry.

Scholars could rather utilise detailed product reviews given by experts to extract design knowledge at various levels of abstraction (e.g., Function-Behaviour-Structure). Extracting such knowledge could be of value to discovering design opportunities and generating problem statements. Domain experts who provide such detailed reviews can identify fundamental issues with a concept that is embodied in the product. An expert mentions all specifications, various use cases, do's/don'ts, estimated lifetime etc. In addition, an expert provides the reviews with necessary context that is often absent in consumer opinions. YouTube provides both expert reviews and consumer opinions on a single platform, which is currently underutilised by scholars.

## 3.6. Other Sources

*3.6.1. Function Structures*

Built upon traditional function structures [254], [255], the functional basis developed by Stone and Wood [152] constitutes functions (e.g., convert, distribute) and flows (e.g., solid material, mechanical energy). The functional basis led to the development of functional models for several products for over 184 electromechanical products and 6906 artefacts [21]. Due to its tremendous popularity, several scholars have attempted to apply and build upon the modelling technique. We review such contributions that are relevant to NLP.

Sridharan and Campbell [256, pp. 141, 143] propose several grammar rules to ensure consistency in functional models. For example, to the function 'remove solid', the secondary inflow – 'mechanical energy' and the outflow – 'reaction force' is added, while, the primary outflow is modified to 'two solids' [256, pp. 145–147]. Sangelkar and McAdams [257] improve on functional models by including user activities obtained from ICF[48] to create action-function diagrams, which they use to compare typical and universal products (e.g., Box Cutter and Fiskars Rotary Cutter).

---

[48] International Classification of Functioning, Disability, and Health - https://tinyurl.com/7ph87kb7

Sen et al. [258] formalize function structures using a prescribed vocabulary for entities and relationships while also proposing several rules for the construction of flows. For example, Rule 14 states [258, p. 6], "A Material flow can have one or more upstream flows, all of which must be of type Material." Agyemang et al. [259] propose several pruning rules to reduce uncertainty and improve consistency in modelling function structures. For example, Rule 8 states [259, p. 504], "Remove all signal, sense, indicate, process, detect, measure, track, and display functions."

To assist with the construction of function structures, Gangopadhyay [260] develop the Augment Transition Network – ATN parser that detects the entities and conceptual dependencies upon providing a text input. To automatically construct functional models, Yamamoto et al. [261] extract (noun, part of, noun) triples (e.g., "wheel of car") using the ESPRESSO algorithm [262] and develop a tree structure, where nouns are replaced by adjacent verbs found in documents.

Wilschut et al. [263, p. 535] extract functions from sentences that comply with a specific grammatical structure; e.g., "Component x provides power p to component y". Using Wikipedia articles on 'machines', Cheong et al [264, pp. 4, 5] obtain and classify Subject-Verb-Object (SVO) triples as functions and energy flows if objects and verbs match with secondary terms in functional basis and their WordNet synonyms. Also, if the combined similarity (Jiang-Conrath and Word2Vec) between the object and 'energy' is greater than 2.9, they classify the object is classified as energy flow.

*3.6.2. Miscellaneous*

We review some purpose-specific classifiers that were built using miscellaneous sources of natural language text. To classify manufacturing concepts using the manufacturing capabilities, Sabbagh et al. [265] label the concepts (e.g., 'annealing', 'hardening') with capabilities (e.g., 'highspeed machining') using the data provided by 260 suppliers listed in ThomasNet[49] and a manufacturing thesaurus [266]. They train the labelled dataset using the following classifiers: Naïve Bayes, k-NN, Random Forest, and SVM. Sabbagh and Ameri [267] obtain LSA-based vectors of manufacturing concepts and cluster these using the manufacturing capability data – ThomasNet for 130 suppliers in heavy machining and complex machining.

---

[49] https://www.thomasnet.com/

To map technical competencies and performances, using the methods such as Probabilistic Latent Semantic Indexing (PLSI), Non-Negative Matrix Factorization (NNMF), and Latent Dirichlet Allocation (LDA), Ball and Lewis [268] extract topics from two corpora: course descriptions and project descriptions of students who were enrolled in the capstone. For each topic and each student, either from course or project, they compute the aggregated score based on his/her grade. They subsequently map course and project vectors using the following methods: Linear Regression, Decision Tree, k-NN, Support Vector Regression, and ANN.

# 4. Discussion

In Section 3, we have reviewed and summarized NLP contributions according to the types of text sources. In the summary sections for each type of text source, we have indicated the method-wise and data-wise limitations, while also mentioning specific opportunities. In this section, we discuss how the NLP contributions thus reviewed could be applied in the design process and what are the potential future directions for the scholars who would contribute to NLP in-and-for design.

## 4.1. Applications

To provide a summary of the design applications that are currently supported by NLP, we utilise the integrated design innovation framework that was developed at the Singapore University of Technology and Design (SUTD). The framework[50] builds upon the double-diamond model of the UK Design Business Council and includes various design modules within each phase. The framework has been utilised to train practitioners from various domains who attend design thinking workshops at the university. Over 20 workshops are held every year – during each workshop (2-3 days), on average, five design innovation facilitators train over 50 practitioners on design thinking. It is important to note that the framework does not span the entirety of the design process, methods, and underlying steps. For instance, the framework does not cover immersed spatial thinking [269]. We utilize this framework to set a boundary for our discussion and to identify the application gaps that could potentially lead to future research opportunities for design scholars.

We list the modules of the design innovation framework across each phase of the design process, as shown in Table 7. For these modules, we highlight the specific steps that are being supported by NLP to indicate the steps that are yet to be supported. We also highlight, on some occasions, the steps as well as NLP applications that could be considered future opportunities. We could consider Table 7 as a minimal NLP guide for choosing a module or a step within a module to develop specific NLP-based supports. In future, as more NLP contributions are reported in the literature, we hope to extend this NLP guide using a comprehensive list of design methods like the Design Exchange[51].

---

[50] https://www.dimodules.com/dilearningmodules
[51] https://www.thedesignexchange.org/design_methods

Table 7: Applications of NLP in the design process. We highlight the currently supported steps within the module and future opportunities.

| Phase | Module | NLP Applications |
|---|---|---|
| **Discover** | **Interviews**: explore usage, identify users, inquire likes/dislikes and use, extract needs and insights | **Sentiment Analysis**: Text Classification, Network Analysis. Topic Modelling, Sentiment Indicators, Clustering, Named Entity Recognition |
| | **Scenarios**: ideate scenarios (how, who, where), present scenarios, observe user reactions | **Product Feature Modelling**: Rule-based Approach, Kano Maps, Regression, Curve Fitting, Clustering, Text Classification, Neural Networks |
| | **User-Journey Map**: gather insights, choose persona, identify touchpoints, identify channels, sketch user journey, rate emotional level, extract opportunities, sketch future journey | **User Profiling**: Clustering, Named Entity Recognition, Affective Attributes, Affective-Design Attribute Relationship <br><br> **Usage Scenarios**: Topic Extraction, Language Models, House-of-Quality, Optimization Methods, Text Classification, Ontology Extraction, FBS, Dependency Parsing, Regression, Kano Map |
| **Define** | **Affinity Diagram**: gather needs, group needs | **Design Rationale Extraction**: Text Classification <br><br> **Emotion Vocabulary**: Clustering |
| | **Personas**: gather persona, consolidate behaviour, present persona | **Kansei Engineering**: Survey, Kansei-Design Matrix, Clustering, Topic Extraction, Multi-Factor Analysis, Kansei-Feature Regression, Text Classification, Term Similarity, Image Classification, Fuzzy Membership Function |
| | **Activity Diagram**: observe user activities, record activities, visualize activity sequence, extract insights | **Action-Function Diagram**: Functional Basis, Rule Mining |
| | **Hierarchy of Purpose**: create opportunity statements, create generalized statements, review statements | **Requirements Elicitation**: Documentation Guidelines, Text Generation, Sentence Completion |
| | **System Functions**: gather needs, map needs and flows generate functions, create function structures | **Functional Modelling**: Grammar Rules, Action-Function Linking, Function Vocabulary, Pruning Rules, Term (Function/Flow) Identification, Ontology Extraction, Text Mining, Text Similarity |
| **Develop** | **Mind Mapping**: initiate design opportunity, generate categories, generate sub-categories, generate solutions, review mind-maps, expand mind maps, reorganize mind maps | **Mind Mapping**: Term Retrieval <br><br> **Patent Mining**: Term-based Patent Map, Class-based Technology Map, Product-based Patent Map, Predicate Logic, Topic Clustering, Phrase Extraction <br><br> **Solution Generation**: TRIZ, Cosine Similarity, Patent Similarity, Function-based Patent Classification |
| | **6-3-5 (C-Sketch)**: form a 6-member group, sketch 3 ideas, pass the sketches to a neighbour, improvise on the sketches, repeat 5 times | **Iterative Labelling**: Object Recognition, Term Retrieval, Image Classification <br><br> **Iterative Annotation**: Text Generation, Sentence Completion, Knowledge Retrieval, Ontology-based Retrieval (e.g., Definitions) <br><br> **Iterative Argumentation**: Text Classification, Text Mining, Clustering, Ontology-based Patent Mining, Knowledge Retrieval |
| | **Design-by-Analogy**: identify keywords, search for inspiration, align relational structure, generate concepts, utilize tools, make inferences, iterate | **Keyword Identification**: Lexical Relationships, Semantic Similarity, SAPPhIRE, FBS, Text Classification, Domain Ontologies <br><br> **Relation-based Retrieval**: Ontological, Lexical, Physical, Ecological, Biological <br><br> **Solution Generation**: Function-based Patent Classification, Text Classification <br><br> **Patent Mining**: FBS-based Term Retrieval, Function-based Patent Similarity, Image Annotation, Ontology-based Topic Modelling |
| | **Real-Win-Worth**: gather solutions, check reality, check novelty, check value | **Concept Association**: Clustering, Topic Association, Network Analysis, Patent Keyword Summarization, Patent Abstract Summarization |
| | | **Novelty Assessment**: Kansei Attributes, Pattern Matching, Semantic Similarity, SAPPhIRE, Topic Modelling |
| **Deliver** | **Multimedia Story Boarding**: identify target user, communicate context, identify key actors, generate flow of events, present story, gather feedback | **User Profiling**: Clustering, Named Entity Recognition, Text Classification, Text Mining, Term Retrieval, Ontology-based Retrieval |
| | | **Usage Scenarios**: Topic Extraction, Language Models, House-of-Quality, Optimization Methods, Text Classification, Ontology Extraction, FBS, Dependency Parsing, Regression, Kano Map |
| | | **Image Annotation**: Entity Recognition, Text Generation, Knowledge Retrieval, Ontology-based Retrieval |
| | | **Kansei Engineering**: Survey, Kansei-Design Matrix, Clustering, Topic Extraction, Multi-Factor Analysis, Kansei-Feature Regression, Text Classification, Term Similarity, Image Classification, Fuzzy Membership Function |
| | **Prototyping Canvas**: choose a solution/concept, fill prototyping canvas, discuss the canvas, build prototype, test prototype, analyse results | **Requirements Elicitation**: Documentation Guidelines, Network Analysis, Text Classification |
| | | **Design Rationale Extraction**: Text Cleaning/Segmentation, Term/Phrase Disambiguation, Text Classification, Text Mining, Clustering, Ontology-based Patent Mining |
| | **Scaled Model**: conduct dimensional analysis, identify key parameters, employ scaling, construct scale model, evaluate model | **Ontology Discovery**: Text Mining, Similarity Measurement, Clustering, Topic Modelling, Patent Similarity |
| | | **Case-based Reasoning**: Knowledge Retrieval, Knowledge Graph Construction, Ontology-based retrieval, Case Indexing |
| | | **Failure Analysis**: Sequence-Sequence Mapping, Text Classification |
| | | **Parameter Identification**: Survey, Kansei-Design Matrix, Kansei-Feature Regression |
| | | **Design Evaluation**: Kansei Attributes, Pattern Matching, Semantic Similarity, SAPPhIRE, Topic Modelling, Text Classification, Success Metrics |

*4.1.1. Discover*

The design innovation framework suggests, as shown in Table 7, that in the *discover* phase, interviews are conducted with potential users to extract needs and insights. Upon collecting user perceptions on specific usage scenarios, a user journey map is developed. Consumer opinions from e-commerce and social media platforms readily provide user profiles along with their ratings, usage, and needs. While the steps in the *discover* modules are largely accomplished using sentiment analysis [182], [194] and usage context extraction [216], [217], Kansei engineering methods capture user emotions for the presented usage scenarios [239], [244].

Kansei engineering also allows establishing a relationship between emotions and product features to predict their importance [247], [248]. The design knowledge thus extracted from consumer opinions is often not sufficient to capture the user journey as the opinions lack enough context and detail. Some seeding information like user persona [207], touchpoints, and channels [208] could be extracted to initialise the user journey map, which could only be developed upon mining detailed expert reviews [215] and conducting user studies [241].

*4.1.2. Define*

In the *define* phase, the user needs are identified and grouped while capturing the user personas to develop activity diagrams. The data generated thus far is utilised to concretize design opportunities and create function structures that map needs to product functions. In terms of gathering needs and persona, the NLP supports remain the same as what was discussed in the *discover* phase. To develop activity diagrams, Sangelkar and McAdams [257] provide partial support by mining association rules from the action-function diagrams.

While there is a need for NLP support in terms of text generation to create opportunity statements, some documentation guidelines have been proposed to structure the requirements such that these are suitable to perform NLP tasks [37], [270]. The scholars have extensively invested in NLP approaches to map needs to functions [89], [140], generate functions [77], [263], and develop function structures [7], [260], [261] as we have reviewed in this article.

*4.1.3. Develop*

The *develop* phase capitalises on the concretised needs, problem statements, and function structures from the *define* phase to generate solutions using various approaches such as mind-map, 6-3-5 sketching, and design-by-analogy. Supports have been developed regarding the mind-maps to generate nodes [94] and organise these into categories [108]. In the absence of user needs, scholars have proposed various approaches to initiate design opportunities from technology maps [160], [162] and biomimicry strategies [143], [271]. The approaches to design opportunity identification could also lend themselves to widening strategies like keyword expansion [83], [145] and concept exploration [99], [100].

While 6-3-5 sketch is often overlooked by scholars in terms of NLP, it is possible to label the sketches using object recognition and image classification algorithms. While the label for such algorithms often tends to be abstract (e.g., man, animal), it is possible to retrieve specific and label-related terms using ontologies and context information [61]. The sketches are often annotated with titles, definitions, the flow of events, etc. To reduce annotation time and make plausible annotations, it is possible to use text generation approaches, especially sentence completion algorithms. In digital sketching interfaces, definitions of components (retrieved from knowledge bases) may pop up on hover.

Scholars have extensively contributed to the research in design-by-analogy in terms of identifying search keywords [82], [86], generating solutions [151], [272], [273], especially via relation-based retrieval algorithms [97], [98]. These supports, however, inform less whether the analogies are suitable. The analogical inferences are therefore yet to be supported.

The design innovation framework shown in Table 7 suggests that the solutions thus generated should be gathered and checked for reality, novelty, and value. The NLP contributions have been effective in associating and discovering categories among several crowdsourced solutions [103], [107]. While several other performance indicators like flexibility and manufacturability are also important metrics to be considered while selecting concepts, computing value is difficult while developing a concept, as value requires sufficient usage context. Current NLP contributions are capable of supporting interim tasks in novelty assessment that is carried out in many ways [274].

*4.1.4. Deliver*

To *deliver* the solutions, the framework suggests creating a multi-media storyboard that communicates the role of solutions in specific scenarios. Scholars have proposed approaches to identify generic users and usage context from consumer opinions that could stimulate ideas for storyboarding. Object-detection algorithms coupled with knowledge graphs [275] could be useful for labelling and describing scenes like storyboards. Kansei engineering methods could be adopted to capture emotional feedback on the storyboards. Besides multimedia storyboarding, NLP techniques could provide direct as well as indirect support for prototyping and developing scaled models.

To build, test, and analyse prototypes, the current NLP supports help understand requirements including dependencies [10], elicit requirements [30], capture design rationale [40], [276], analyse failures [39], [64], and facilitate case-based reasoning [60], [61]. While these existing supports are applicable for testing scaled models as well, building a scaled model requires dimensional analysis that maps the key design parameters (e.g., viscosity) onto the performance parameters (e.g., energy consumption).

As an alternative to dimensional analysis, scholars have adopted deep learning approaches to associate design and performance parameters. For example, upon combining three datasets[52], Robinson et al. [277] map building feature like area, the number of floors, heating degree days, building activity etc., onto the annual energy consumption using several models like gradient boost, multi-layer perceptron, KNN, SVR etc [277, p. 894]. While performance parameters like energy consumption are largely derived from industry standards, the influential design parameters could be chosen and evaluated based on Kansei methods [239], [241].

## 4.2. Methodological Directions

Based on our review, we propose eight methodological directions for future NLP applications to support the design process.

First, we prioritize the extraction of knowledge graphs from text, which will be utilised in the design process as a knowledge base. Second, we recommend the development and utilisation of domain-specific language models to perform tasks such as classification, NER, and question-answering. In the third and fourth directions, we propose the development of one or more text generation and neural machine translation models. Next, we propose the adoption of NER methods and collaborative tagging approaches to facilitate the tasks such as classification, relation extraction etc. Further, we propose that scholars develop standard datasets using design text as a common evaluation platform for future NLP applications. Finally, we propose to develop success metrics for evaluating the efficacy of NLP supports.

We have listed these directions along with examples in Table 8. We provide specific examples for the first six directions using a publicly available text[53]. For the remainder of this section, we explain these directions in individual sub-sections.

---

[52] Commercial Buildings Energy Consumption Survey - https://www.eia.gov/consumption/commercial/
New York City Local Law 84 (LL84) - https://www1.nyc.gov/html/gbee/html/plan/ll84_about.shtml
New York City Primary Land Use Tax Lot Output (PLUTO) - https://www1.nyc.gov/site/planning/data-maps/open-data/dwn-pluto-mappluto.page
[53] https://ceramicpro.com/what-does-a-nano-ceramic-coating-do/

Table 8: Methodological directions with examples.

| Methodological Direction | Example Input | Example Output |
|---|---|---|
| **Design Knowledge Graph** | "A nano-ceramic coating, a scientifically formulated solution meant to penetrate microscopic imperfections, fill those gaps in the top range of the nanoscale and provide a layer of protection that's nearly as strong as solid quartz. 9H ceramic coating work by bonding with the existing surface to form a protective nano-ceramic shield on the surface." | <nano-ceramic coating, penetrate, microscopic imperfections><br><nano-ceramic coating, provide, protection layer><br><nano-ceramic coating, form, nano-ceramic shield><br><nano-ceramic coating, bond, existing, surface><br><existing surface, bond, nano-ceramic shield> |
| **Domain-Specific Language Model** | "Nanoceramic coating", "solution", "ceramic" | Embeddings |
| **Text Generation** | "Nano-ceramic coating" | "Nano-ceramic coating provides protection layer"<br>"Nano-ceramic coating forms nano-ceramic shield"<br>… |
| **Neural Machine Translation Model** | "Nano-ceramic coating provides protection layer"<br>"Nano-ceramic coating forms nano-ceramic shield" | <nano ceramic coating, $hasFunction$, provide protection layer><br><nano ceramic coating, $hasBehaviour$, form nano ceramic shield> |
| **Named-Entity Recognition** | "A *nano-ceramic coating*, a scientifically formulated solution meant to penetrate microscopic imperfections, fill those gaps in the top range of the nanoscale, and provide a layer of protection that's nearly as strong as *solid quartz*" | "Nano-ceramic coating" – coating material, coating solution<br>"Solid quartz" – coating material, coating solution |
| **Collaborative Tagging** | "A nano-ceramic coating, a scientifically formulated solution *meant to* penetrate microscopic imperfections, fill those gaps in the top range of the nanoscale and *provide* a layer of protection that's nearly as strong as solid quartz. 9H ceramic coating *work* by *bonding* with the existing surface to *form* a protective nano-ceramic shield on the surface." | "A nano ceramic coating…" – Function<br>"9H ceramic coating…" – Behaviour |
| **Standard Datasets** | **NLP Tasks**: Text Classification, Text Similarity…<br>**NLP Applications**: Functional Representation, Design Rationale Extraction… | Standard Datasets |
| **Success Metrics** | Text Comprehension, Keyword Diversity, Problem Understanding… | Success Metrics |

*4.2.1. Design Knowledge Graph*

A knowledge graph comprises facts of the form – $\{\langle h, r, t \rangle\}$ and serves as an infrastructure for the development of various NLP applications. A design knowledge graph includes facts like <'stapler', 'comprises', 'leaf spring'>, < 'hammer', 'push', 'staple'> that could be utilised or generated in the design process. A design knowledge graph carries informative as well as reasoning advantages over networks [278] that provide pairwise statistical [79], semantic [133], and syntactic [78] relationships among a large collection of design terms (lexicon).

To process and recognize entities in text sources like internal reports, design concepts, and consumer opinions [279], it is necessary to build design knowledge graphs that could replace the common-sense lexicon

(e.g., WordNet). Domain-specific ontologies (e.g., QuenchML) capture design knowledge using relationships ($r$) like 'hasProperty', 'partOf', and 'hasWeight' that are technically preferable in comparison with that of common-sense ontologies like ConceptNet, i.e., the relationships such as 'atLocation' and 'usedFor' captured by these. However, domain-specific ontologies capture abstractions (e.g., <Component, hasWeight, xx>) rather than facts (e.g., <clamp, weighs, 65 grams>) that are extracted from natural language text and captured using knowledge graphs.

We have shown an example in Table 8 for the facts that could be extracted from a sample text. As discussed in Section 3.4.4, technical publications that include patents and scientific articles are preferable sources for extracting facts and developing design knowledge graphs due to their high accessibility, information content, and quality. Scholars have indicated the possibility of extracting triples from the patent text [280]–[282]. Siddharth et al. [283], for example, apply some rules to extract facts from patent claims by exploiting the syntactic and lexical properties. While patents could offer rule-based extraction methods due to consistent language, scientific articles require a mix of rule-based, ontology-based, and supervised approaches.

*4.2.2. Domain-Specific Language Model*

Early models of language given by traditional grammar have often proposed a restricted set of rules for forming sentences [284, pp. 5, 6], which limits the opportunity to produce a vast number of sentences. The modern view of a language model involves training large corpora to capture the likelihood of a given sequence of words (or tokens), e.g., "metallic bond is strong" in the same order. Originally developed as N-gram models, these models have evolved into deep learning-based models or transformers such as BERT and GPT-x. These models advance the theory of acquisition model of a language [284, p. 38] via statistical embeddings of generative grammar, which is otherwise represented as Parts of Speech, Structural Dependencies etc.

These models capture the embeddings of tokens and sequences through masked language modelling where a large number of sequence-sequence pairs are provided as training data. The input-output pair must belong to the same sequence but nearly 15% of the input tokens are expected to be masked. The embeddings that result from these models could be directly used to train classifiers, sequence-to-sequence tasks like Q & A, and NER tasks. Several variants of BERT have been introduced at the corpora level, e.g., BioBERT [75] and at the architecture level, e.g., k-BERT [285]. The variant k-BERT, for instance, stitches facts from a domain knowledge graph onto the tokens for training the model.

Using domain-knowledge embedded language models like k-BERT provides embeddings of terms that are meaningful. As opposed to common techniques like BOW, LSA, and Word2Vec, embeddings from domain-specific language models should return 'nearly true' cosine similarity between a pair of artefacts (described using text) that have domain-association, similar physical properties, and perform similar functions. Moreover, such domain-specific embeddings could aid in efficient concept retrieval in the respective domain. For example, a radiology-specific language model should identify the terms closest to "Magnetic Resonance Imaging" than a common-sense language model.

*4.2.3. Text Generation*

Originally referred to as Natural Language Generation (NLG) systems, e.g., the DOCSY model [286], applications that generate text reduce cost, ensure consistency, and maintain the standard of documentation [287, pp. 261–265]. Such applications are relevant to the design process where requirements must be elicited, opportunity statements must be generated, and solutions must be described. In Table 8, we indicate an example where a seeding term "nano-ceramic coating" results in plausible sentences using text generation algorithms.

To support ontology-based verification of requirements, Moitra et al. [270, p. 347] propose that a requirement shall be expressed as follows: REQUIREMENT R (name); SYSTEM shall set $x$ of $X$ to $x_1$ (conclusion); when $y \in Y$ (condition). Likewise, scholars have proposed templates for describing design concepts as well [106], [157], [288]. While such a template-based approach works with a limited scope, it is necessary to implement text generation algorithms that are built out of RNNs, LSTM, and Transformers.

Zhu and Luo [289] fine-tune GPT-2 for mapping the problems (including categories) to solutions using problem-solution data obtained from RedDot[54]. They also explore the capabilities of GPT-3 that support analogy-by-design in terms of generating text descriptions upon providing source-target domain labels as inputs. For a given technology domain, using KeyBERT[55], Zhu and Luo [290] extract topics (terms and keyphrases) from patent titles and create a dataset of topic-title pairs. They fine-tune GPT-2 for mapping topics to titles so that solutions (as hypothetical titles) could be generated using search keywords (topics of interest).

---

[54] https://www.red-dot.org/

[55] https://github.com/MaartenGr/KeyBERT

*4.2.4. Neural Machine Translation*

Neural Machine Translation (NMT) models are trained to map sequence-to-sequence using an encoder-decoder framework [291]. These models are often associated with Transformers owing to the similarity in structure and behaviour of the neural networks that were built to accomplish the mapping task. NMT models have been specifically built to perform cross-language translation tasks and these are useful to increase semantic interoperability in design environments. For example, the rules "Smith Ltd shares machines with NZ-based companies" and "Smith Ltd allows NZ-based companies to use its machines" mean the same but are written in different forms [70].

In Table 8, we have shown an example of semantic forms that could be mapped from design text through neural machine translation. To standardize manufacturing rules, Ye and Lu [70] map a manufacturing rule into a semantic rule using a neural machine translation model [292] that comprises an encoder and a decoder with 256 Gated-Recurrent Units (GRUs) present in each [70]. Chen et al. [293] propose semantic rule templates to formalize requirements so that these are easily verified using ontologies. NMT models coupled with semantic rule templates are necessary to translate ambiguous natural language sentences into a machine-readable form.

*4.2.5. Named Entity Recognition*

NER is a sequence-to-sequence task like POS tagging where entities and their respective tags are identified, e.g., 'General Electric' as an organisation and 'San Francisco' as a location. From a design perspective, the term 'fan' shall be recognized as a product and the terms 'ceiling fan', 'exhaust fan', and 'CPU cooling fan' shall be recognized with specific categories. While plenty of NER models and associated datasets exist for common-sense entity recognition [294], design-based datasets and models are yet to evolve. NER is also the first step towards the extraction of knowledge graphs, as described in Section 4.2.1.

In Table 8, we have shown that in a given design text, entities like "nano-ceramic coating", and "solid quartz" must be identified using tags like coating material, and coating solution etc., Before the identification of entity tags, it is necessary to recognize terms that comprise one or more words (n-grams). Scholars have often utilised POS tags, dependencies, and ontologies to recognize n-grams. Due to poor performance, such approaches must be replaced with deep-learning models, as demonstrated by Chiarello et al. [144] in their NER application.

*4.2.6. Collaborative Tagging System*

Collaborative tagging (or folksonomy) is useful for the classification of a large set of documents as well as sentences in these. This bottom-up approach has been recently popular instead of a traditional top-down approach where the classification scheme is defined by the experts, e.g., International Patent Classification. The current classifications in vast knowledge sources like Patent Databases, Web of Science, and Encyclopaedia are less useful for developing NLP applications to support the design process. For instance, the classification codes that are assigned to a patent could inform the type of invention but not its purpose, behaviour, and components.

We have indicated an example in Table 8 for the design-specific tags that could be assigned to individual sentences in a text document. The tags shall be recommended based on external knowledge as well as the previous tags [295]. These tags could also be expanded using classifiers [27]. While several advantages to collaborative tagging exist, scholars are yet to introduce or develop many interfaces that help to assign tags to documents that are universally accessible. COIN platform is an example of such a collaborative tagging system [296]. The use of such interfaces in design education, workshops, and laboratory settings allows a variety of tags to be assigned to an open-source document that could be reused for developing retrieval algorithms.

*4.2.7. Standard Datasets*

None of the NLP contributions that we have reviewed in this article leverage a design-specific gold standard dataset for evaluation. If an embedding technique is used for measuring the similarity between text descriptions of two artefacts, what is the trueness of that similarity? Similarly, if an application combines several tasks like NER, classification etc., to extract FBS from text, what is the efficacy of the application? For such cases, scholars are currently creating their datasets from scratch, which reduces the possibility of comparing different applications within design research.

A gold standard dataset is necessary for NLP applications that aim to measure artefact level metrics such as novelty, feasibility etc. These metrics shall be measured in different ways, but it is recommended that scholars provide a gold standard for different ways to benefit the development of NLP applications. For example, given a text description of an artefact, a dataset may include the novelty scores measured using distance-based and frequency-based approaches while also indicating the reference product databases utilised for the measurement.

*4.2.8. Success Metrics*

A variety of NLP applications have been and will be developed to support various design tasks. To ensure the efficacy of these applications, success metrics are necessary. The metrics like accuracy for classification only tell us that the classifier performs well on the test data. However, the utility of such a classifier is often assessed based on the artefact level metrics such as novelty, quantity, variety etc. While such metrics are crucial, it would be useful to also measure the 'goodness' of envisioned scenarios, activity diagrams, mind maps, opportunity statements, search keywords, requirement formulation etc.

The expert designers spend a majority of the time proposing and evaluating solution alternatives [297, p. 430], while novices spend more time understanding the problem. Even if novices generate quick solutions, experts have a better ability to recognize good solutions. Novices could therefore significantly benefit from NLP support in terms of keyword recommendation, opportunity statements, identifying novel solutions etc. Since novices need to develop expertise throughout the design process, success metrics at each step could be beneficial for their learning as well as for understanding the efficacy of NLP supports.

## 4.3. Theoretical Directions

While the proposed methodological directions could impact the development of NLP applications in the near future, our review also led us to raise a few questions regarding constructs that embody the design-centric natural language text and the roles of these constructs in the design process. Addressing these questions could be of importance in the extended future to facilitate the development of cognitive assistants that make independent decisions in the design process based on long-term memory and extensive reasoning capabilities. We discuss these questions in the remainder of this section.

*4.3.1. Characteristics and Constructs*

In our review, we have indicated the text characteristics of various types of natural language text sources that are utilised or generated in the design process. These characteristics are only relevant to the NLP methodologies applied to the text sources. The literature does not communicate the characteristics of natural language that allow us to distinguish a piece of text that is relevant to the process. Let us consider the following sentences for example.

1) "The pan is heated while the steak gets seasoned,"
2) "During the recrystallization stage, the material is heated above its recrystallization temperature, but below its melting temperature."

The first sentence mentions a cooking tip and the second one is part of the annealing process[56]. The underlying factors of distinguishability between these two sentences are unclear. If we assume that the distinction could be attributed to the usage of technical ('recrystallization', 'temperature', 'material') and common-sense ('pan' and 'steak') terms, it is also possible that these terms could be used interchangeably in other text sources. Hence, we raise the first open question as follows.

*What are the unique characteristics of natural language text that are relevant to the design process?*

While the efforts to identify the design-specific characteristics in natural language may lead to a bifurcation of technical and common-sense natural language text, it is necessary to acknowledge that design knowledge is present in various flavours within the common-sense text as well. We provide an example using the reviews of a Scotch-Brite kitchen wiper on Amazon[57].

- Affordance – "I am using this not in kitchen but as a car wash assessary to clean all windows…"
- Recommendation – "You can definitely buy this product…"
- Satisfaction – "The quality of this one is ok"
- Feature description – "…the green color rubber part is very small and thin"
- Characterization – "I am not sure about this durability"
- Aesthetics – "Too small and badly designed"
- Technical description – "…the actual size of the blade is mere 6.2 inches, which is too small for cleaning a large surface area… the blade is bent at an angle of almost 30-40° to the handle…"

From our review, we are unable to obtain sufficient explanation for the assignment and evolvement of the knowledge categories that we have tied to the sentences in the above example. Scholars have conducted large-scale analyses on consumer opinions while informing a little on what these sources communicate in the context of design. The constructs of design knowledge that embody the natural language text are often captured by ontologies and language models. These systems, however, are not capable of providing a cogent explanation of the phenomenon behind the judgement of design knowledge in a given text. It is therefore important to understand the following.

---

[56] https://www.metalsupermarkets.com/what-is-annealing/
[57] t.ly/1ifJ

*What are the unique constructs that embody design knowledge into natural language text?*

There has been extant literature on ontologies that have aimed to address the question above. These ontologies are built by domain experts (top-down) as well as extracted from text sources (bottom-up). The outcomes of these approaches have often been distinguishable [296]. In addition, there exists a significant difference in the level of abstraction between elementary [3], [44] and abstract ontologies [298], [299].

Despite the recent attempts to extract abstract ontologies from text, e.g., SAPPhIRE [300], and FBS [77], it is easier to recognize elementary ontologies, as indicated by various knowledge retrieval systems developed using these. The elementary ontologies, however, do not cover a large scope of design like abstract ontologies. To address the above-mentioned question, it is, therefore, necessary to obtain investigate the following.

*How to bridge elementary and abstract ontologies to support the design process?*

*4.3.2. Comprehension*

The following questions relate to the performances of the natural language text concerning comprehension in the design process. Let us consider a natural language explanation for the firing cycle of a Glock handgun[58].

"… when the trigger is pulled, this pulls the firing pin backward … a connector pin that guides the connector in a downward motion… this motion frees up the firing pin, allowing it to strike the primer…"

While the above-stated text captures components and the causality of events, it is hard to visualise the orientations and positions of components like 'trigger', 'firing pin', and 'connector pin' without (annotated-) images. In addition, the text is only pertinent to the firing cycle and does not include other subsystems of the handgun like the safety mechanism. It is difficult to interpret and reproduce the knowledge of system architecture (the hierarchy of a handgun in this example) purely using natural language text. Hence, a multi-modal explanation is often necessary, especially in the design process [288]. The affordance in comprehension through textual mode shall therefore be investigated as follows.

---

[58] https://ghostinc.com/ghost-inc-blog/how-does-a-glock-work/

*What is the expected level of comprehension offered by natural language text in the design process?*

Addressing the above-stated question could set a boundary for the performance of NLP applications. Large-scale analyses on crowdsourced natural language text (e.g., consumer opinions) often seem to highlight the lack of information quality, while providing less importance to the amount of design knowledge offered within a particular window of text. Since consumer opinions must adhere to word limits on platforms like Amazon and Twitter, usage scenarios are often captured through images and videos. It would be worth investigating how text could be elaborated such that it provides a level of comprehension similar to that of a multi-modal explanation. We, therefore, ask the following question.

*How to elaborate natural language text to obtain the desired level of comprehension in the design process?*

*4.3.3. Creativity*

Cognitive scientists define an insight or 'Aha' moment as the instance of sudden realization that is often associated with a stimulus. In terms of semantic memory, insight occurs when there is a new connection between entities that lead to a sequence of new connections [301]. Such insights are necessary for solving problems, especially during the design process. A particular case of insight occurs in the design process when there is a relational alignment between two pairs of entities [302].

Let us consider an example. Wall-climbing robots adopt various adhesive mechanisms to establish contact with the climbing surface. These mechanisms are less effective when robots are heavy and the surfaces are hard, flat, and smooth. Let us consider a stimulus for this design problem. Mudskippers climb slippery surfaces of rocks by generating a vacuum at the limb interface. The interaction – 'generate vacuum' at the rock interface fits well in the wall interface. Such an alignment of relation creates an insight that leads to 'making sense of new interactions like releasing vacuum, decreasing pressure etc.

Scholars have proposed representation schemes like FBS and SAPPhIRE to model 'far' domain examples such that relations are explicitly shown. A majority of far-domain examples, however, are only available as natural language text that often does not explicitly state these relations. It is difficult to surf through several documents to encounter such relations and experience insights. To address this issue, scholars have proposed to summarize several documents by extracting the representative terms [157]. These terms alone are insufficient for gathering insights due to the lack of context. It is therefore necessary to investigate the following question.

*How to represent natural language text such that design insights are maximised?*

Souza et al. [13] generate short summaries of patent documents through an LSTM-based sequence-to-sequence mapping. Such statistical approaches are less guided by design theories that inform the constructs of design knowledge that should be present in such summaries. While a succinct representation of natural language text is necessary for gathering insights, it is also important to form the right queries to search for documents that could potentially include stimuli for solving design problems.

The mental representation of a design problem is translated to opportunity statements that are simplified into search keywords that form queries. It is a common phenomenon that the search results often guide the development of more keywords. If the initial set of keywords is not representative of the problem statement, the user has the chance to be misled by the results. In such situations, an expert could provide reliable guidance on how the problem statement is translated into search keywords by identifying the gaps and discrepancies in the problem formulation.

Let us consider an example of pumping water out of the basement. A direct search for terms like "pumping water", "basement water" etc. might lead to several unwanted results. An expert, on the other hand, might question the type of basement, the cause of water in the basement, the type of water, the basement surroundings etc. These intricate details help elaborate the problem statement, from which the expert could extract important cues and translate these into keywords that are appropriate as well as technical (if necessary). It is therefore worth examining how problems should be narrated such that it is possible to translate these into meaningful opportunity statements and in turn appropriate search keywords.

From our review, we understand that keyword expansion approaches are largely driven by the search results alone [15], [86] rather than by the missing details of the problem statement. The current NLP applications are therefore less capable of playing the expert's role in examining the problem statement. To address this caveat, it is necessary that scholars provide a theoretical explanation to the following question.

*How to narrate a design problem such that it is better translated to appropriate search keywords?*

We expect that in the future, NLP applications recommend keywords that are guided by the problem statement and provide results using succinct natural language text such that more insights are experienced in the design process. Given that insights often lead to solutions to design problems in the form of design concept alternatives, it is necessary to choose among these alternatives for implementation and testing purposes. Several design metrics like feasibility, novelty, utility etc., are being used to choose the alternatives.

Given that human judgement on alternatives often involves extensive effort and bias, scholars have proposed some NLP applications to compute the design metrics using natural language text data [41], [113]. Herein, both the alternatives and reference material (e.g., Kansei attributes) comprise natural language text. Since the usage of terms in the text descriptions of concept alternatives significantly impacts the judgement of design metrics it is important to address the following question.

*What is the role of natural language in the judgement of design metrics?*

## 4.4. Summary

From our review of 223 articles related to NLP in-and-for design research, we identified the supported applications in the design process using a framework as discussed in Section 4.1. We have also indicated the steps and modules within the framework that are currently not supported by NLP. While we expect that such gaps are addressed by scholars in the near future, we hope that an NLP guide is developed using a more comprehensive design framework. We expect that such a guide informs the following for an individual module: type of text sources used/generated, example case studies, relevant state-of-the-art NLP methods, rubrics to evaluate NLP methods etc. After summarizing the applications, we presented the directions (listed in Table 9) for the advancement of NLP in-and-for design.

Table 9: Summary of methodological and theoretical directions

|   | **Methodological Directions** |
|---|---|
| 1. | Design Knowledge Graph – Text Cleaning, Term Identification, Relation Extraction, Functional Representation, Question Answering, Graph-Based Reasoning, Graph Embedding etc. |
| 2. | Domain-Specific Language Model – Text Classification, Named Entity Recognition, Sentence Completion, Sentiment Analysis, Term Extraction, Similarity Measurement etc. |
| 3. | Text Generation – Sentence Completion, Requirements Elicitation, Statement Generation, Technical Documentation etc. |
| 4. | Neural Machine Translation – Sentence Disambiguation, Storage Compression, Language Standardisation etc. |
| 5. | Named Entity Recognition – Term Identification, Ontology Construction, Term Disambiguation, Document Indexing, Knowledge Graph Extraction, Functional Representation etc. |
| 6. | Collaborative Tagging – Text Classification, Document Indexing, Ontology Construction, Sentiment Detection etc. |
| 7. | Standard Datasets – Text Classification, Creativity Assessment, Functional Representation etc. |
| 8. | Success Metrics – Text Comprehension, Problem Diversification, Problem Detailing, Solution Assessment, Keyword Expansion etc. |
|   | **Theoretical Directions** |
| 1. | What are the unique characteristics of natural language text that are relevant to the design process? (Characteristics) |
| 2. | What are the unique constructs that embody design knowledge into natural language text? (Constructs) |
| 3. | How to bridge elementary and abstract ontologies to support the design process? (Constructs) |
| 4. | What is the expected level of comprehension offered by natural language text in the design process? (Comprehension) |
| 5. | How to elaborate natural language text to obtain the desired level of comprehension in the design process? (Comprehension) |
| 6. | How to represent natural language text such that design insights are maximised? (Creativity) |
| 7. | How to narrate a design problem such that it is better translated to appropriate search keywords? (Creativity) |
| 8. | What is the role of natural language in the judgement of design metrics? (Creativity) |

The methodological directions are necessary to enhance the performances and conduct a robust evaluation of NLP applications in-and-for design. In Table 9, we have also indicated the downstream tasks and applications that could entail the methodological directions. While design knowledge bases, text generation, and named entity recognition could be developed using state-of-the-art NLP approaches, language models and neural machine translation require further improvement in core NLP. For the remaining methodological directions, scholars may consider operationalising the existing design theories into metrics and datasets so that NLP applications could be developed without theoretical challenges.

The theoretical directions call for an understanding of the characteristics and constructs of natural language text that influence the affordance of comprehension and creativity in the design process. As the volume of natural language text data grows multi-fold with time, it is necessary to distinguish the text that is applicable to the design process. The characteristics and constructs that constitute design language should also indicate the missing elements of design knowledge that influence the abilities to form search keywords, comprehend design text, generate insights, and judge the solutions.

The proposed directions primarily call for an understanding of the structure and role of the design language that should help bolster the performances of natural language text in learning, design, and computational environments. For example, in a computational environment, a piece of text (e.g., a movie review) may return an accurate sentiment score. In another example, a well-written chapter on kinematics may be useful in a learning environment. These two examples, however, may be less useful in a design environment. Similarly, a design text (e.g., technical requirement) may perform poorly in learning and computational environments. In order not to be misled by the performance in a single environment, it is important to distinguish natural language text by identifying the characteristics and constructs that constitute design language.

# Conclusions

The purpose of this review article was to encapsulate a large body of NLP contributions that are relevant to the design process so as to identify unsupported design applications, potential methodological advancements, and gaps in design theory. We gathered 223 articles published in 32 journals for our review. We organise, explain, and examine these articles according to the type of text sources: internal reports, design concepts, discourse transcripts, technical publications, and consumer opinions. We then discuss our findings in terms of design applications and future directions. The overall conclusions from the review and the entailing discussions are as follows.

1. A comprehensive NLP guide is necessary for the identification of specific design modules and developing NLP supports according to the type of text sources utilised/generated in these.
2. While several methodological directions could be pursued using state-of-the-art NLP tools, the development of standard datasets and success metrics require the operationalisation of existing design theories.
3. It is necessary to identify the unique characteristics and constructs that help distinguish design-centric natural language text as well as influence the performances in terms of comprehension and creativity in the design process.

# APPENDIX I

We use the Web of Science[59] advanced search to retrieve the articles for review. We input all queries in the following format,

$$((TS = kw1^*OR\ kw2^* \ OR\ kw3^*\ldots)\ OR$$

$$(TI = kw1^*OR\ kw2^* \ OR\ kw3^*\ldots)\ OR$$

$$(AB = kw1^*OR\ kw2^* \ OR\ kw3^*\ldots))\ AND$$

$$(SO = dj1\ OR\ dj2\ldots)$$

where TS = Topic/keyword, TI = Title, AB = Abstract, SO = Journal, kw ∈ {keyword list}, and dj ∈ {journal list}.

We executed the queries on 19th September 2021 and the outcomes of each query are shown in Table 1A.

Table 1A: Precisions of different queries.

| # | Query Step | Results | Relevant | Precision % |
|---|---|---|---|---|
| 1 | Keywords, Design journals | 890 | 95 | 10.674 |
| 2 | *Expanded keywords*, Design journals | 1744 | 102 | 5.849 |
| 3 | Expanded keywords, *Expanded Design journals* * | 2328 | 117 | 5.026 |
| 4 | Expanded keywords, *Web of Science* | 6930765 | 223 | 0.003 |
| 5 | Expanded keywords, Web of Science, *article type* | 4908353 | 223 | 0.005 |
| 6 | *Expanded keywords including 'design'*, Web of Science, article type | 593765 | 206 | 0.035 |
| 7 | Expanded keywords including 'design', Web of Science, article type, *selected categories* | 78919 | 206 | 0.261 |
| 8 | Expanded keywords including 'design', *selected journals, journals with count >= 10*, article type, and selected categories * | 6523 | 206 | 3.158 |

* Denotes the step where we manually read the titles, abstracts, and full texts to obtain the final set of articles

We explain the queries as shown in Table 1A for the remainder of this section. In the first query, we consider eight 'well-known' design journals[60] using the following keywords: 'semantic', 'text', 'language', 'pars', 'ontolog', 'abstract', 'word', 'phras', 'sentence'. We retrieve 890 articles and obtain the frequent terms from topics (> 1), titles (> 4), and abstracts (> 4) to identify more keywords – 'vocabular', 'sentiment', 'gramma', 'lexic', 'linguistic', 'syntactic', and 'term'. We include these additional keywords in the second query to retrieve 1,744 articles. To include more journals that fall within the scope of design research, we consult the literature that provides a broad view of design research [303], [304] as well as reviews [305]. Based on the literature, we include five additional journals[61] in the third query to retrieve 2,328 articles.

---

[59] https://mjl.clarivate.com/search-results

[60] We included the following design journals: Artificial Intelligence for Engineering Design Analysis and Manufacturing, Research in Engineering Design, Journal of Engineering Design, Design Studies, Design Science, Journal of Mechanical Design, Journal of Computing and Information Science in Engineering, International Journal of Design.

Since NLP applications that benefit design research could also be published outside the design journals, in the fourth query we remove the journal filter and retrieve 6,930,765 results. Since these results also include conference proceedings and book chapters, in the fifth query, we select only journal articles to retrieve 4,908,353 articles. To filter these, in the seventh query, we include an additional keyword 'design' and particular subject categories[62] to retrieve 78,919 articles. For these articles, we manually selected the journals using the following criteria: article count $\geq$ 10, non-distant domain (e.g., not "Journal of Biological Chemistry"), non-specific topic (e.g., not "Applied Surface Science"), general design-related (e.g., Computers in Industry), technology-related (e.g., Scientometrics). These filters result in 6,523 articles.

We merge the results of the third and final queries as the first three queries did not include the 'design' keyword filter. We examine the titles and abstracts[63] of the merged results to obtain 277 articles. Upon reading the full texts of 277 articles, we obtain the final set – 223 articles that we have made accessible on Github[64]. Using the final set of articles, we also report the precisions of each query as shown in Table 1A.

---

[61] We include the following journals in addition to the previously populated list of design journals: Design Journal, Design Quarterly, Design Issues, International Journal of Design Creativity and Innovation, Journal of Computer-Aided Design.

[62] The selected categories include the following: Engineering Manufacturing, Engineering Multidisciplinary, Engineering Mechanical, Computer Science Interdisciplinary Applications, Computer Science Software Engineering, Art, Computer Science Artificial Intelligence, Engineering Industrial, Social Sciences Interdisciplinary, Architecture.

[63] An example of discarded result had the following text within the abstract: "… material properties that are computed in terms of the microstructural texture descriptors." [306].

[64] https://github.com/siddharthl93/nlp_review/blob/4b9e6b378c8df0bbf61a36e466a50dbb5a0a65d2/nlp_review_papers.csv